\documentclass[10pt,twocolumn,letterpaper]{article}

\usepackage{cvpr}              % To produce the CAMERA-READY version
\usepackage{amsmath}
\usepackage{amssymb}
\usepackage{amsthm}
\usepackage{booktabs}
\usepackage{enumitem}
\usepackage{graphicx}
\usepackage{makecell}
\usepackage{mathtools}
\usepackage{marvosym}
\usepackage{microtype}
\usepackage{multicol}
\usepackage{multirow}
\usepackage{ragged2e}
\usepackage{subcaption}
\usepackage{tabularx}
\usepackage[table,dvipsnames]{xcolor}
\usepackage{tikz}
\usepackage{url}
\usepackage{wrapfig}
\usepackage{xspace}
\usepackage{threeparttable}

\usepackage[pagebackref=false,colorlinks,citecolor=magenta,urlcolor=magenta, linkcolor=magenta]{hyperref}

\usepackage[capitalize]{cleveref}
\crefname{section}{Sec.}{Secs.}
\Crefname{section}{Section}{Sections}
\Crefname{table}{Table}{Tables}
\crefname{table}{Tab.}{Tabs.}

\definecolor{forestgreen}{rgb}{0.133, 0.545, 0.133}
\definecolor{yellowyellow}{rgb}{0.133, 0.545, 0.133}

\definecolor{correct}{RGB}{173, 173, 173}
\definecolor{incorrect}{RGB}{192, 0, 0}

\definecolor{forestgreen}{rgb}{0.133, 0.545, 0.133}
\definecolor{yellowyellow}{rgb}{0.984, 0.714, 0.071}

\def\paperID{37096} % *** Enter the Paper ID here
\def\confName{CVPR}
\def\confYear{2026}

\title{LiDAR Prompted Spatio-Temporal Multi-View Stereo for Autonomous Driving}

\author{
Qihao Sun\textsuperscript{\rm 1,2,$\dag$} \quad
Jiarun Liu\textsuperscript{\rm 1,$\dag$} \quad
Ziqian Ni\textsuperscript{\rm 1,$\dag$} \quad
Jianyun Xu\textsuperscript{\rm 1} \quad
\\
Tao Xie\textsuperscript{\rm 2} \quad
Lijun Zhao\textsuperscript{\rm 2} \quad
Ruifeng Li\textsuperscript{\rm 2,\Letter} \quad
Sheng Yang\textsuperscript{\rm 1,\Letter} \\
\textsuperscript{\rm 1}Unmanned Vehicle Dept., CaiNiao Inc., Alibaba Group, \quad 
\textsuperscript{\rm 2}Harbin Institute of Technology \\
{\tt\small 23s008047@stu.hit.edu.cn, shengyang93fs@gmail.com}
}

\begin{document}

\maketitle

\newcommand{\unnumberedfootnote}[1]{%
\begingroup
\renewcommand{\thefootnote}{}%
\footnotetext{#1}%
\addtocounter{footnote}{0}%
\endgroup
}
\unnumberedfootnote{$\dag$: Equally contributed. \Letter: Corresponding author.}

\begin{abstract}
Accurate metric depth is critical for autonomous driving perception and simulation, yet current approaches struggle to achieve high metric accuracy, multi-view and temporal consistency, and cross-domain generalization.
To address these challenges, we present \textbf{DriveMVS}, a novel multi-view stereo framework that reconciles these competing objectives through two key insights: 
 (1) Sparse but metrically accurate LiDAR observations can serve as geometric prompts to anchor depth estimation in absolute scale, and (2) deep fusion of diverse cues is essential for resolving ambiguities and enhancing robustness, while a spatio-temporal decoder ensures consistency across frames.
Built upon these principles, DriveMVS embeds the LiDAR prompt in two ways: as a hard geometric prior that anchors the cost volume, and as soft feature-wise guidance fused by a triple-cue combiner.
Regarding temporal consistency, DriveMVS employs a spatio-temporal decoder that jointly leverages geometric cues from the MVS cost volume and temporal context from neighboring frames. 
Experiments show that DriveMVS achieves state-of-the-art performance on multiple benchmarks, excelling in metric accuracy, temporal stability, and zero-shot cross-domain transfer, demonstrating its practical value for scalable, reliable autonomous driving systems.
Code: \url{https://github.com/Akina2001/DriveMVS.git}.
\end{abstract}
 
\section{Introduction} \label{sec:intro}

Nowadays, large-scale crowd-sourced data collected by robotaxis across diverse road and driving conditions enables rigorous validation and iterative improvement of perception systems through generative reconstruction~\cite{shao2025chronodepth,hu2025depthcrafter,du2025rgegs} or world modeling~\cite{liang2025seeing,karypidis2024DinoForesight,zeng2025sss}.
Within this closed-loop simulation framework, effective spatial modeling from casually captured real-world driving clips becomes a prerequisite for preserving physical realism in the learned world representation.
As 3D vision systems evolve under commercial pressure to reduce sensor costs, modern L4 autonomous vehicles are increasingly adopting minimalist LiDAR configurations -- using fewer sensors to balance safety, redundancy, and cost-effectiveness.
Therefore, it's essential to build a robust depth estimation pipeline that leverages reliable, high-fidelity 3D metrics.

% Recent advances in spatial modeling have established three dominant paradigms for zero-shot depth estimation:
Existing depth estimation approaches fall into three broad families, each with a characteristic limitation in driving scenes:
(1) Monocular foundation models (\eg, DepthAnything~\citep{yang2024depthanything, yang2024depthanythingv2}, MoGe-2~\citep{wang2025moge}), which leverage large-scale pretraining and temporal cues for strong cross-domain generalization and efficient inference, yet suffer from scale ambiguity and limited temporal consistency;
(2) General-purpose MVS models (\eg, MVSAnywhere~\citep{izquierdo2025mvsanywhere}), which combine monocular priors with multi-view geometry for high-fidelity reconstructions but typically estimate depth independently per frame -- leading to temporal flickering -- and degrade under low parallax, static motion, or texture repetition situations due to unreliable epipolar cues;
(3) Feed-forward multi-view models (\eg, VGGT~\citep{wang2025vggt}, MapAnything~\citep{keetha2025mapanything}), which enable fast, end-to-end prediction but perform inferior in absolute depth accuracy.
While multi-modal fusion methods~\citep{lin2025promptda, wang2025priorda} mitigate some issues by anchoring depth to sparse LiDAR data, these prompts are inherently sparse, intermittent, and unevenly distributed due to occlusions and sensor limitations. Systems that rely solely on current-frame cues become fragile when inputs are missing or degraded, leading to distorted 3D structures and inaccurate scene recovery.

These challenges lead to a unifying insight: for reliable deployment in real-world autonomous driving, a depth estimation system must simultaneously satisfy four key requirements under typical minimalist LiDAR configurations: 
(1) Metric-scale accuracy, even when multi-view cues fail due to low parallax, static motion, or textureless regions, by maintaining persistent metric anchoring through sparse prompts;
(2) Temporal consistency, achieved through explicit modeling of temporal context to ensure smooth, flicker-free predictions across sequences;
(3) Robustness to prompt intermittency and mild misalignment, with no degradation when LiDAR inputs are partial or absent;
(4) Zero-shot cross-domain generalization, preserving the broad applicability of foundation models across diverse, unseen environments.

Guided by this principle, we propose \textbf{DriveMVS}, a novel MVS framework that simultaneously achieves metric-scale accuracy, temporal consistency, robustness to prompt dropout in the zero-shot setting, and cross-domain generalization.
The core of it is a metric-embedding design that directly integrates sparse prompts into cost-volume construction by explicitly disentangling relative consistency learning from absolute scale anchoring.
Besides, our Triple-Cues Combiner employs a novel Transformer-based strategy to intelligently fuse these anchored geometric cues with powerful structural priors and high-fidelity sparse metric guidance.
Finally, our Spatio-Temporal Decoder, enhanced with a motion-aware temporal layer, ensures smooth, stable, and metrically accurate depth propagation across video sequences.
% \todo{
% a triple-cue combiner that provides soft, feature-wise guidance by adaptively weighting monocular, multi-view, and prompt features. 
% For temporal consistency, a motion-aware spatio-temporal decoder jointly exploits geometry from the MVS cost volume and context from neighboring frames, enabling smoothing and scale propagation when direct prompts are missing. 
% Robustness is promoted by stochastic prompt dropout during training and by cross-view propagation of metric scale at inference, enabling the system to maintain fidelity when LiDAR coverage is sparse or absent. }
Extensive experiments on KITTI~\citep{geiger2012kitti}, DDAD~\citep{packnet2020ddad}, and Waymo~\citep{sun2020waymo} show that DriveMVS delivers high-fidelity, metrically accurate, and temporally stable depth across challenging conditions, consistently outperforming prior state-of-the-art methods. 
We also demonstrate that DriveMVS effectively transfers its learned capabilities across different datasets and domains.

Our contributions are as follows:
\begin{itemize}
    \item We present \textbf{DriveMVS}, an MVS pipeline which unifies absolute scale accuracy, cross-domain generalization, and robust temporal consistency. It effectively addresses the limitations of existing approaches by integrating sparse metric guidance and spatio-temporal reasoning.
    
    \item We design a metric embedding mechanism that explicitly anchors geometric cues to absolute scale and intelligently fuses them with structural priors and high-fidelity metric prompts to resolve ambiguity precisely.

    \item Experiments demonstrate that DriveMVS achieves state-of-the-art performance across multiple challenging autonomous driving benchmarks. Its superior performance in metric accuracy, temporal stability, and overall robustness against sensor and environmental variations demonstrates the practical value for scalable, reliable real-world applications.
\end{itemize}

\section{Related Work} \label{sec:related}

\paragraph{Monocular Depth Estimation Models.}
Monocular depth estimation (MDE) has evolved from hand-crafted cues~\citep{hoiem2007recovering,saxena2008make3d} to deep, data-driven approaches~\citep{eigen2014depth,guo2025murre,fu2018deep} that markedly improve accuracy but struggle out-of-domain.
Recent work pursues strong zero-shot generalization by scaling data and training signals: the DepthAnything series~\citep{eigen2014depth, yang2024depthanythingv2, chen2025videodepthanything}; additional advances include diverse scene priors~\citep{xian2018monocular, xian2020structure}, affine-invariant losses~\citep{ranftl2020towards}, and transformer architectures~\citep{ranftl2021dpt}.
Diffusion-based MDE opens a complementary path: Marigold adapts image diffusion priors for depth~\citep{ke2024marigold,rombach2022stablediffusion}; DepthLab sharpens structure with depth guided generation~\citep{liu2024depthlab}; flow matching accelerates sampling~\citep{gui2025depthfm}; and video diffusion enables open world video depth~\citep{hu2025depthcrafter}.
While these approaches recover compelling relative geometry, they typically inherit the scale ambiguity of generative models and thus fail to recover metric-consistent depth.
To address this problem, classic methods rely on RGB-D or LiDAR data in specific domains (\eg, indoor scenes or street views)~\citep{bhat2021adabins,yin2019enforcing}, while more recent efforts~\citep{butler2012naturalistic, guizilini2023towards, kendall2018multi} demonstrate improved cross-domain generalization. 
Complementary to these image-only advances, practical deployments often pair cameras with LiDAR, motivating the use of sparse depth as a geometric prior for dense, metrically calibrated maps~\citep{schonberger2016pixelwise,jensen2014large}.
Recently, OMNI-DC~\citep{zuo2024omnidc} models varying sparsity with probability-based losses and normalization.
Prompt conditioned approaches adapt foundation models or impose explicit geometric constraints under scarce priors, as in PromptDA and PriorDA~\citep{lin2025promptda,wang2025priorda}.
These approaches show that sparse priors can effectively guide dense prediction within monocular frameworks, while leaving multi-view and spatiotemporal modeling to complementary components.

\paragraph{Multi-View Feed-Forward Models.}
Feed-forward models replace iterative optimization with single-pass inference while keeping explicit geometric outputs.
DUSt3R~\citep{wang2024dust3r} and MASt3R~\citep{leroy2024mast3r} predict coupled pointmaps from which cameras, poses, and dense geometry are recovered post hoc; follow-ups~\citep{elflein2025light3rsfm, pataki2025mpsfm, murai2025mast3rslam, duisterhof2025mast3rsfm} integrate these predictions into classical SfM and SLAM pipelines for large-scale reconstruction.
To move beyond two-view coupling, transformer models with latent state memory (Spann3R~\citep{wang2025spann3r}, CUT3R~\citep{wang2025cut3r}, MUSt3R~\citep{cabon2025must3r}) maintain a persistent scene representation and enable direct multi-view reconstruction without external bundle adjustment.
Multi-view variants extend two-view backbones: MV-DUSt3R+~\citep{tang2025mv_dust3r_plus} parallelizes cross-attention to support multiple references, VGGT~\citep{wang2025vggt} uses alternating attention to jointly predict pointmaps, depth, pose, and tracking features.
$\pi^3$~\cite{wang2025pi3} fine-tunes VGGT to decouple reference coordinates.
MapAnything~\citep{keetha2025mapanything} adopts a fully factored representation that separates per-view ray directions and along-ray depth from global camera poses and a single scene scale, which improves heterogeneous input support, calibration flexibility, and seamless batching for offline processing.
Together, these trends push feed-forward models toward unified, memory-efficient multi-view reasoning with better scalability and throughput.

\paragraph{Multi-View Stereo Models.}
Multi-view stereo (MVS) recovers \emph{metric} depth under calibrated cameras by enforcing epipolar constraints.
Traditional MVS includes voxel-based~\citep{osman2017semantic,kutulakos2000theory,seitz1999photorealistic}, point cloud-based~\citep{furukawa2009accurate, lhuillier2005quasi}, and depth map-based~\citep{xu2020planar, xu2022multi} methods, with depth map-based methods dominating because they decouple per-view depth estimation from fusion.
However, handcrafted matching remains brittle under illumination changes, low-texture surfaces, and non-Lambertian surfaces~\citep{schonberger2016pixelwise,furukawa2015mvs}.
Learning-based MVS replaces handcrafted similarity measures with deep features; MVSNet~\citep{yao2018mvsnet} formulates MVS as feature extraction, cost-volume construction via differentiable homography, and cost-volume regularization, inspiring advances in long-range context and geometry~\citep{cao2022mvsformer,cao2024mvsformer++}, explicit handling of occlusions and dynamics~\citep{wimbauer2021monorec,long2020occlusion}, and efficiency~\citep{sayed2022simplerecon,yu2020fastmvsnet} through recurrent and coarse-to-fine design.
Diffusion-assisted MVS (\eg, Murre~\citep{guo2025murre}) introduces SfM priors to enhance cross-view consistency. 
Foundation-style systems such as MVSAnywhere~\citep{izquierdo2025mvsanywhere} emphasize cross-domain and cross-range robustness without test-time retraining. 
Yet MVS degrades in low-parallax or repetitive-texture regimes where epipolar cues are weakened.
% \sqh{
% To address this, recent literature has explored prior-guided architectures.
% For instance, \citep{song2023prior} utilizes external prior depth maps to explicitly constrain the multi-view stereo search space, while \citep{cheng2024stereo} propagates temporal context directly across frames to ensure smooth video depth.
% Furthermore, \citep{bartolomei2024vpp4dc} demonstrates that casting depth completion as a stereo matching problem significantly enhances cross-domain generalization.
% These concepts align with a broader trend in the literature that bridges depth completion, active sensor fusion, and advanced deep stereo matching paradigms to achieve robust, cross-domain geometry estimation (\eg, \citep{lipson2021raftstereo, poggi2022mvgmvs, xin2023simplemapping, conti2024DepthOnDemand} and related surveys like \citep{tosi2025survey}).
% These advances motivate our hybrid formulation that directly combines metric priors with MVS cues to couple global-scale awareness with local geometric consistency.
% }

\section{Methodology}
\label{sec:method}

\begin{figure*}[t]
    \begin{center}
    \includegraphics[width=1.0\textwidth]{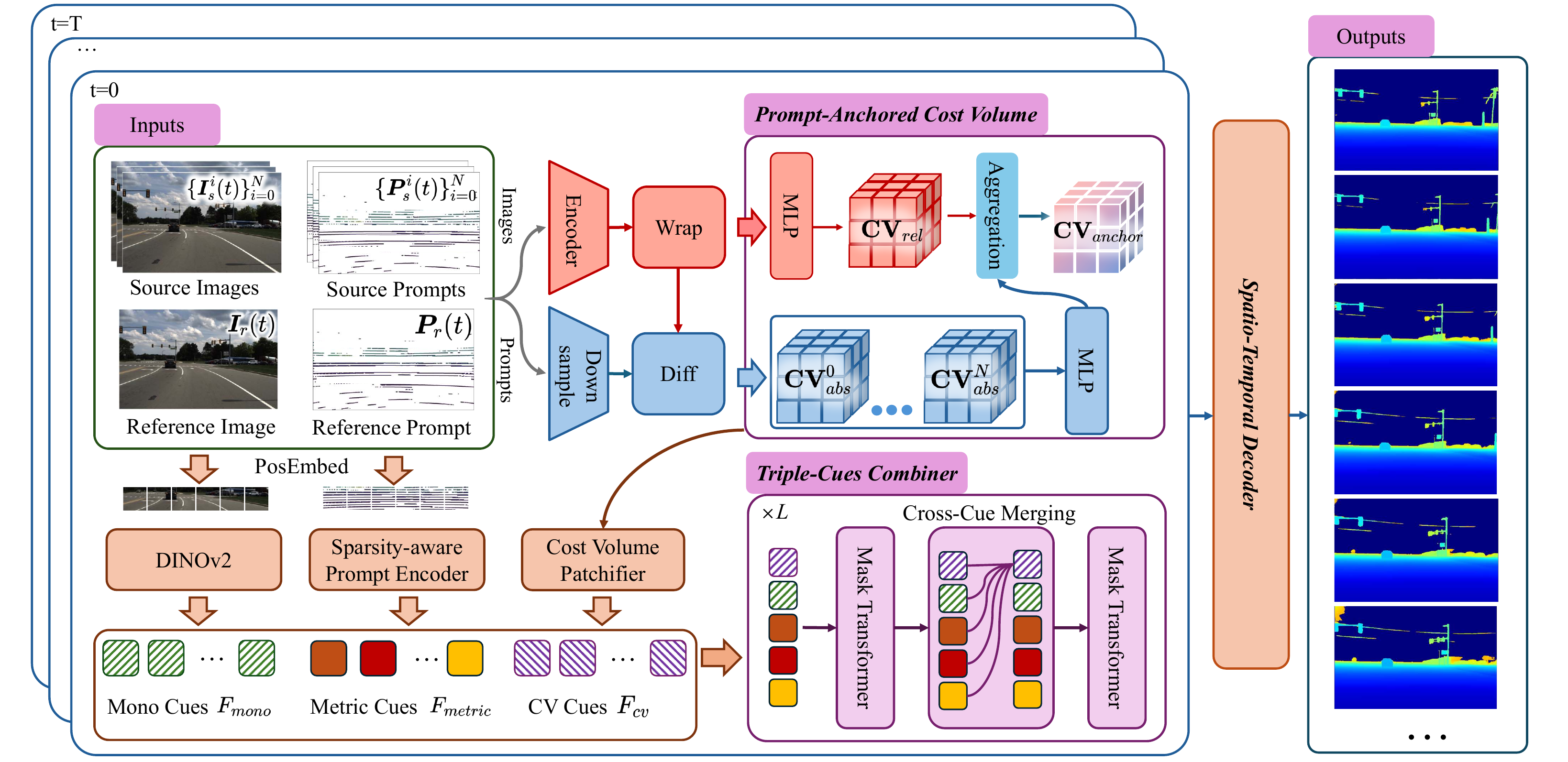}
    \end{center}
    \vspace{-0.2cm}
    \caption{
    \textbf{Overview of MVS-Pro.} We introduce a Prompt-Anchored Cost Volume (\cref{sec:prompt_anchored_cv}) mechanism to fuse the absolute depth metric prompt into the multi-view cost volume. To fuse heterogeneous features from images, depth, and cost volume, we propose a Triple-Cues Combiner (\cref{sec:triple_cues_combiner}) to combine the cues. Finally, a Spatio-Temporal Decoder (\cref{sec:spatio_temporal_decoder}) produces continuous, consistent depth results. 
    }
\label{fig:system}
\end{figure*}

%%添加Doubletake和simplerecon的引用

\subsection{Problem Definition}
\label{sec:problem}

Our model operates on a sequence of length $T$.
At each time step $t\in[0,T)$, the input $\boldsymbol X_t$ consists of the following components:
a reference image $\boldsymbol I_r(t)\in \mathbb{R}^{H\times W\times 3}$;
a set of $N$ source images $\{\boldsymbol I_s^i(t)\}_{i=0}^N$;
the corresponding intrinsics and extrinsics for all $N+1$ views;
and the corresponding sparse metric prompts $\boldsymbol P(t)\in\mathbb{R}^{H\times W}$ for all views.
Our model $\mathcal{M}_{\theta}$ maps the full sequence $\{\boldsymbol X_t\}_{t=0}^T$ to a corresponding sequence of per-pixel logit maps $\{\boldsymbol x(t)\}_{t=0}^T$, where $\boldsymbol x(t) \in \mathbb{R}^{H \times W}$.
\begin{equation}
    \mathcal{M}_{\theta}(\{\boldsymbol X_t\}_{t=0}^T) = \{\boldsymbol x(t)\}_{t=0}^T.
\end{equation}
After that, the absolute metric depth $\hat{\boldsymbol D}(t)\in\mathbb{R}^{H\times W}$ is calculated with the help of the cost volume, which will be discussed in the following sections.

\subsection{Prompt-Anchored Cost Volume}
\label{sec:prompt_anchored_cv}
Our pipeline builds upon the view-count agnostic cost volume mechanism~\cite{izquierdo2025mvsanywhere, sayed2022simplerecon}.
Given a reference image $\boldsymbol I_r(t)$ and source images $\{\boldsymbol I_s^{i}(t)\}_{i=0}^{N}$, deep features $F_r$ and $\{F_s^{i}\}_{i=1}^{N}$ at $\frac{H}{4} \times \frac{W}{4}$ resolution are extracted through the first two stages of ResNet-18~\cite{he2016resnet}.
For each reference pixel $(u_r, v_r)$ and each of $\mathcal{D}$ depth hypothesis planes $k$ ($\mathcal{D}=64$ bins, sampled uniformly in log space), it re-projects into all $N$ views to assemble a per-view metadata that encodes relative geometric cues (\eg, feature dot products ($F_r \cdot F_s^i$), ray directions, relative poses, and validity masks).
A single MLP then maps this metadata to a per-view score; softmax across views yields aggregation weights, and the plane cost is a weighted sum.
However, this design learns almost entirely from relative consistency (feature matching, multi-view geometry). In low-parallax motion (\eg, traffic jams) or in textureless regions, these cues become ambiguous, often collapsing metric scale and reducing performance to scale-ambiguous monocular estimation.

To address this issue, we introduce the Prompt-Anchored Cost Volume (PACV), which explicitly disentangles this process and separately learns relative consistency and absolute scale anchoring by different MLPs, as illustrated in~\cref{fig:system}.
Concurrently, we compute the same metadata as above and feed it to an MLP to obtain an intermediate per-view feature $\mathbf{CV}_{rel}(k,j)$, representing the learned \textit{relative consistency cost volume} for plane $k$ and source view $j$.
For the current depth hypothesis $d_k$ and the $N+1$ downsampled sparse prompts $\boldsymbol P_{r,s}$, we build an \textit{absolute metric cost volume} by taking their absolute differences at depth $d_k$, using a masking value ($-1$) for invalid pixels, following \citep{sayed2024doubletake}.
We concatenate these costs across views and pass them through a lightweight MLP to produce an intermediate absolute feature $\mathbf{CV}_{abs}(k,j)$.
Finally, we aggregate them (\eg, concatenation) to form a unified anchored feature:
\begin{equation}
    \phi(k,j) = \mathrm{Concat}(\mathbf{CV}_{rel}(k,j),\mathbf{CV}_{abs}(k,j)).
\end{equation}

The \textit{anchored cost} $\mathrm{CV}_{anchor}\in\mathbb{R}^{\mathcal{D}\times\frac{H}{4}\times\frac{W}{4}}$ is computed as following:
\begin{equation}
\begin{aligned}
    & \omega(k,j), s(k,j) = \mathrm{MLP}(\phi(k,j)), \\
    & \mathbf{CV}_{anchor}(k) = \Sigma_j\left[\mathrm{Softmax}(\omega(k,j))\odot s(k,j)\right], \\
    & \mathbf{CV}_{anchor} = \mathrm{Concat}(\mathrm{CV}_{anchor}(k)),
\end{aligned}
\end{equation}
where $\omega(k,j), s(k,j)$ are the weights and scores decoded from an MLP, and $\mathbf{CV}_{anchor}$ is the softmax-weighted sum of these scores.

By enforcing the network to jointly reason over learned relative consistency ($\mathbf{CV}_{rel}$) and explicit absolute metric cues from prompts ($\mathbf{CV}_{abs}$) prior to score prediction and weight assignment, the PACV prevents cost volume collapse in low-parallax or textureless regions where relative cues alone are ambiguous or unreliable under geometric constraints.

\subsection{Triple-Cues Combiner}
\label{sec:triple_cues_combiner}
In addition to PACV, we introduce the Triple-Cue Combiner (TCC), a transformer-based aggregation mechanism that reasons jointly over three heterogeneous cue streams with complementary properties:
\begin{itemize}
\item \textbf{CV Cues} ($F_{{cv}}$): dense cues produced by the Cost Volume Patchifier~\cite{izquierdo2025mvsanywhere}, which encode per-depth hypotheses and are geometrically anchored but structurally agnostic.
\item \textbf{Mono Cues} ($F_{{mono}}$): cues from a DINOv2 encoder initialized with Depth-Anything-V2 weights, which provide strong global context and a scene-level \emph{relative} depth prior.
\item \textbf{Metric Cues} ($F_{{metric}}$): sparse cues from sparsity-aware prompt encoder, which supply high-fidelity \emph{absolute} metric constraints.
\end{itemize}

As illustrated in~\cref{fig:system}, TCC is a deep, $L$-layer Mask Transformer (we set $L=12$). Each basic block comprises two Mask Transformer units, separated by a Cross-Cue Merging module that enables the structured fusion of heterogeneous cues.
The first and last stages of each basic block are implemented as a Mask Transformer, which allows each cue to independently refine its own internal, non-local representations.
It consists of three parallel self attention ($\text{SA}$) operations, masked from each other:
\begin{equation}
\begin{aligned}
    & F'_{{cv}} = \text{SA}(F_{{cv}})+F_{{cv}} \\
    & F'_{{mono}} = \text{SA}(F_{{mono}})+F_{{mono}} \\
    & F'_{{metric}} = \text{Mask-SA}(F_{{metric}})+F_{{metric}},
\end{aligned}
\end{equation}
where $\text{Mask-SA}$ denotes standard self-attention with an explicit mask that prevents queries from attending to tokens corresponding to invalid or missing prompt pixels. This masking ensures robustness to sparsity and avoids erroneous propagation of unreliable signals.
% % 在supp中补充可视化结果
The second stage, Cross-Cue Merging, performs the core fusion operation.
After channel-wise alignment, we element-wise sum the depth-sensitive, geometrically anchored features $F'_\mathrm{cv}$ with the monocular features $F'_\mathrm{mono}$, which encode strong relative-depth priors:
\begin{equation}
Z = F'_{cv}\oplus F'_{mono}.
\end{equation}
This lightweight summation achieves token-level consistency while maintaining computational efficiency.
Next, the fused representation $Z$ interacts with the metric cue $F'_{metric}$ via cross attention ($\text{CA}$), where $Z$ serves as the query and $F'_{metric}$ provides both keys and values:
\begin{equation}
    \hat{F}_{{cv}} = Z + \text{CA}(\text{Q}=Z, \text{K}=\text{V}=F'_{{metric}}).
\end{equation}
Crucially, this interaction is restricted to valid prompt locations within a spatio-temporal neighborhood spanning multiple frames, ensuring temporal coherence and local fidelity.
Besides, cross-frame prompts further provide temporal consistency and robustness to transient missing cues.

\subsection{Spatio-Temporal Decoder}
\label{sec:spatio_temporal_decoder}
Built on DPT~\cite{ranftl2021dpt}, our decoder upsamples to full resolution while embedding motion-aware temporal self-attention within the upsampling blocks.
It jointly treats the fused tokens and adjacent reference frames, yielding a smooth, stable video depth and enabling scale propagation.
It aggregates information along the temporal dimension via a \emph{temporal layer} comprising a Multi-Head Self-Attention (MSA) module and a Feed-Forward Network (FFN).
Crucially, to enable the model to capture pose changes across frames, we introduce a Relative Pose Encoder that embeds relative camera poses into the feature stream prior to the temporal attention mechanism.
This explicit integration of geometric motion context allows the temporal layer to more effectively discern pixel correspondences and motion.
When feeding features $F_i$ into the temporal layer, self-attention is performed exclusively along the temporal axis to facilitate interactions among temporal features. 
To capture inter-frame positional relationships, we encode absolute positional embeddings over the video sequence. 
The spatio-temporal decoder uniformly samples 4 feature maps from $F_i$ as input. 
To manage computational overhead, temporal layers are strategically inserted only at a few lower-resolution stages.
Finally, absolute metric depth $\hat{\boldsymbol D}(t)\in\mathbb{R}^{H\times W}$ is then recovered by rescaling the normalized output of the sigmoid function $\sigma$ to match the absolute metric scale.
This follows the formulation:
\begin{equation}
\hat{\boldsymbol D}(t) = \exp(\log(d_{\min}) + \log(d_{\max} / d_{\min}) \cdot \sigma(\boldsymbol x(t))),
\end{equation}
where $d_{\min}$ and $d_{\max}$ represent the absolute metric bounds of the cost volume.

\subsection{Implementation Details}
\label{sec:implementation}
% \sqh{In this section, we provide essential information about the network design, prompt normalization, training loss, training data, and training strategy.}

% \todo{\paragraph{Network details.}
% % ViT-B
% We utilize the ViT-Base model as our reference image encoder.
% % sparsity-aware promtp encoder
% % As mentioned above, the triple-cues combiner need to 整合metric cues from sparsity-aware prompt encoder.
% % 具体而言，对于稀疏prompt，首先通过浅层卷积网络进行
% $P\in\mathbb{R}^{H\times W \times 1}$
% % The sparsity-aware prompt encoder 由一个shallow convolutional network (comprises two convolutional layers with a kernel size of 3 and a stride of 1)

% More details can be found in the supplementary materials.} \jr{Depend on the length of the paper. Can be put in the supplementary.}

\paragraph{Prompt normalization.}
During the sparsity-aware prompt encoder, raw sparse LiDAR prompts are transformed by a four-stage sparsity-aware CNN into $1/16$-resolution masked metric features $F_{metric}$, making them feature-space compatible and spatially aligned with the network's prediction space.
However, the irregular range of prompt input in autonomous driving scenarios can impede network convergence.
To address this, we project the sparse metric depths into logit space by first min-max normalizing the logarithm of depth to the bounds of the cost volume and then applying the logit transform:
\begin{equation}
\begin{aligned}
    &d_{norm} = \frac{\log(d_{metric} - \log(d_{\min})}{\log(d_{\max})-\log(d_{\min})}, \\
    &d_{target} = \text{logit}(d_{norm})=\log(\frac{d_{norm}}{1-d_{norm}}).
\end{aligned}   
\end{equation}
This procedure ensures scale consistency between the sparse depth prompts and the network's outputs, facilitating faster, more stable convergence during training.

\paragraph{Training loss.}
% Our total loss function $\mathcal{L}$ is a weighted sum of a spatial loss $\mathcal{L}_{spatial}$ for single-frame geometric accuracy and a temporal loss $\mathcal{L}_{temporal}$ for inter-frame consistency.
To ensure precise per-frame geometry, we adopt the supervised losses from \cite{izquierdo2025mvsanywhere}, which comprise an L1 loss between the log of the ground truth and the log of the predicted depth values $\mathcal{L}_{depth}$, a gradient loss $\mathcal{L}_{grad}$, and a normals loss $\mathcal{L}_{normals}$. 
To enforce temporal stability, we employ the temporal loss $\mathcal{L}_{temporal}$ from \cite{chen2025videodepthanything}, which penalizes inconsistencies in depth changes between consecutive frames.
Training losses are applied to four output scales of the decoder. 
Our final loss is
\begin{equation}
    \mathcal{L} = \alpha(\mathcal{L}_{depth} + \mathcal{L}_{grad} + \mathcal{L}_{normals}) + \beta(\mathcal{L}_{temporal}),
\label{eq:total-loss}
\end{equation}
where $\alpha = \beta = 1$. Please check our \textit{supplementary materials} for detailed loss design and implementation.

\paragraph{Training data.}
\begin{table}
\centering
\caption{We train on four MVS datasets from a variety of domains.
    All these datasets are synthetically rendered, providing perfect ground-truth depths and camera calibration.
    All the sparse prompts are synthesized directly from the ground-truth depths.
    }
\resizebox{1\columnwidth}{!}{%
\begin{tabular}{lllllll}
\toprule
\textbf{Split} & \textbf{Name} & \textbf{Scenes}	& \textbf{\# Total}	& \textbf{\# Total} & \textbf{Sparse}	\\
& & & \textbf{scenes}	& \textbf{images} & \textbf{prompt} \\
\midrule\midrule		
\multirow{4}{*}{\rotatebox{90}{\textit{Training}}}
& TartanAir~\citep{wang2020tartanair} & Indoor, Outdoor & 347	& ~1M & \multirow{4}{*}{Synthetic} \\
% & TartanAirV2~\citep{wang2020tartanair} & Indoor, Outdoor & 1129 & ~1M & \\
& TartanGround~\citep{patel2025tartanground} & Indoor, Outdoor & 789 & ~1M & \\
& VKITTI2~\citep{cabon2020vkitti2,gaidon2016vkitti}	& Outdoor, Driving & 50 & ~21K & \\
& MVS-Synth~\citep{huang2018deepmvs}	& Outdoor, Driving & 117	& ~12K & \\
% & KITTI-CARLA~\citep{deschaud2021kitticarla} & Outdoor, Driving & - & - & \\
\midrule
\multirow{3}{*}{\rotatebox{90}{\textit{Testing}}}
& KITTI~\citep{geiger2012kitti} & Outdoor, Driving & 61 & ~42K & \multirow{3}{*}{LiDAR} \\
& DDAD~\citep{packnet2020ddad} & Outdoor, Driving & 200 & ~16K & \\
& Waymo~\citep{sun2020waymo} & Outdoor, Driving & 202 & ~39K & \\
\bottomrule
    \end{tabular}
    }

    \label{tab:training-datasets}
\end{table}

To enable DriveMVS to generalize across domains, we train on a large, diverse set of synthetic datasets, as listed in \cref{tab:training-datasets}. At the same time, accurate depth supervision of synthesized data can also reduce the introduction of noise during training, helping the model output more accurate depth contours, which indirectly improves the model's zero-shot ability~\cite{zuo2024omnidc}. 
We simulate LiDAR point projections in the camera coordinate system by back-projecting accurate, dense ground-truth depth using randomized extrinsic parameters and ray-sampling patterns. 
To further approximate real-world measurement noise, we introduce outliers and boundary perturbations during sampling, effectively corrupting the prior with realistic artifacts such as missing returns, spurious hits, and edge distortions. Please see our \textit{supplementary materials} for more details.

\paragraph{Training strategy.}
Specifically, we randomly drop each prior modality with a probability of $0.5$ during training. 
When a modality is dropped, its corresponding tokens are set to zero.
This simple yet effective strategy enhances the model's robustness by encouraging the network to learn resilient representations under partial input conditions, enabling minimal degradation when certain priors are unavailable during inference. It yields a single, unified model that can flexibly adapt to various combinations of available priors, eliminating the need for multiple specialized variants.

\section{Experiments}
\label{sec:exp}
\subsection{Experiment Setups}
\paragraph{Benchmark and Baselines.} 

\begin{table*}[ht]
  \centering
  \begin{threeparttable}
    \caption{Benchmark study of depth estimation methods. All methods are evaluated on the official validation splits of \textit{KITTI}~\cite{geiger2012kitti}, \textit{DDAD}~\cite{packnet2020ddad}, and \textit{Waymo}~\cite{sun2020waymo}. Each dataset reports three metrics. MAE is measured in meters (m), while AbsRel and $\tau$ are reported as percentages (\%). The \textbf{best} and \underline{second best} scores are highlighted in \textbf{bold} and \underline{underline}.}
    \label{tab:benchmark}
    
    \small 
    \setlength{\tabcolsep}{4pt} 

    \begin{tabular}{l l ccc ccc ccc}
    \toprule
    \multirow{2}{*}{\textbf{Method}} & \multirow{2}{*}{\textbf{Venue}} & \multicolumn{3}{c}{\textbf{KITTI}} & \multicolumn{3}{c}{\textbf{DDAD}} & \multicolumn{3}{c}{\textbf{Waymo}}
    \\
    \cmidrule(lr){3-5} \cmidrule(lr){6-8} \cmidrule(lr){9-11}
    & & \textbf{MAE}($\downarrow$) & \textbf{AbsRel}($\downarrow$) & \textbf{$\tau$}($\uparrow$) & \textbf{MAE}($\downarrow$) & \textbf{AbsRel}($\downarrow$) & \textbf{$\tau$}($\uparrow$) & \textbf{MAE}($\downarrow$) & \textbf{AbsRel}($\downarrow$) & \textbf{$\tau$}($\uparrow$)
    \\
    \midrule 
    \multicolumn{11}{l}{\textbf{a) Feed-forward reconstruction}}
    \\
    VGGT~\citep{wang2025vggt} & CVPR'25 & $13.19$ & $77.58$ & $2.21$ & $31.30$ & $98.17$ & $0.00$ & - & - & - \\
    MapAnything~\citep{keetha2025mapanything} & arXiv'25 & $1.45$ & $10.14$ & $94.45$ & $11.05$ & $40.57$ & $2.82$ & - & - & - \\
    MapAnything\tnote{\textdagger}~\citep{keetha2025mapanything} & arXiv'25 & $1.27$ & $8.45$ & $95.28$ & $4.60$ & $13.57$ & $81.06$ & - & - & - \\
    \midrule
    \multicolumn{11}{l}{\textbf{b) Monocular depth (w/o prompts)}}
    \\
    % DAv2-L~\citep{yang2024depthanythingv2} & NIPS'24 & $10.45$ & $62.51$ & $3.21$ & - & - & - & - & - & - \\
    MoGe-2~\citep{wang2025moge} & CVPR'25 & $3.60$ & $23.18$ & $34.23$ & $9.23$ & $26.33$ & $27.05$ & $4.97$ & $20.51$ & $50.66$ \\
    DepthPro~\citep{bochkovskii2024depthpro} & ICLR'25 & $2.50$ & $15.44$ & $80.71$ & $11.95$ & $33.65$ & $34.14$ & $7.99$ & $33.07$ & $32.16$ \\
    \midrule
    \multicolumn{11}{l}{\textbf{c) Monocular depth (w/ prompts)}}
    \\
    % DepthLab~\citep{liu2024depthlab} & arXiv'24 & $0.921$ & $2.171$ & $4.498$ & $8.379$ & $5.425$ & $5.275$ \\
    % Marigold-DC~\citep{viola2024marigolddc} & ICCV'25 & $0.558$ & $1.676$ & $2.985$ & $7.905$ & $1.772$ & $4.791$ \\
    % DMD3C~\citep{} & CVPR'25 \\
    PromptDA~\citep{lin2025promptda} & CVPR'25 & $2.40$ & $8.91$ & $87.19$ & $9.02$ & $25.62$ & $67.31$ & $9.00$ & $39.24$ & $44.35$ \\
    Marigold-DC~\cite{viola2025marigolddc} \scriptsize{bfloat16} & ICCV'25 & $0.86$ & $5.43$ & $97.23$ & $3.80$ & $13.83$ & $87.96$ & - & - & -  \\
    PriorDA~\citep{wang2025priorda} & arXiv'25 & \underline{$0.61$} & \underline{$2.98$} & \underline{$98.57$} & \underline{$2.79$} & \underline{$5.82$} & \underline{$94.50$} & \underline{$1.80$} & \underline{$6.04$} & $85.55$ \\
    \midrule
    \multicolumn{11}{l}{\textbf{d) MVS-based depth (w/o prompts)}}
    \\
    MVSFormer++~\cite{cao2024mvsformer++} & ICLR'24 & $3.94$ & $28.50$ & $85.79$ & $5.52$ & $18.20$ & $83.90$ & $10.45$ & $49.41$ & $35.50$ \\
    MVSAnywhere~\citep{izquierdo2025mvsanywhere} & CVPR'25 & $1.78$ & $10.48$ & $90.91$ & $4.18$ & $10.16$ & $91.71$ & $3.30$ & $11.43$ & \underline{$89.80$} \\
    \midrule
    \multicolumn{11}{l}{\textbf{e) MVS-based depth (w/ prompts)}}
    \\
    % MVSAnywhere~\citep{izquierdo2025mvsanywhere} \scriptsize{w/ Doubletake} & CVPR'25 & $1.78$ & $10.48$ & $90.91$ & $4.18$ & $10.16$ & $91.71$ & $3.30$ & $11.43$ & \underline{$89.80$} \\
    % MVSAnywhere~\citep{izquierdo2025mvsanywhere} \scriptsize{w/ mvgMVS} & CVPR'25 & $1.78$ & $10.48$ & $90.91$ & $4.18$ & $10.16$ & $91.71$ & $3.30$ & $11.43$ & \underline{$89.80$} \\
    \cellcolor{violet!7}\textcolor{violet!66}{\textbf{Ours}} & \cellcolor{violet!7}\textbf{-} & \cellcolor{violet!7}\textbf{0.49} & \cellcolor{violet!7}\textbf{2.56} & \cellcolor{violet!7}\textbf{98.78} & \cellcolor{violet!7}\textbf{2.64} & \cellcolor{violet!7}\textbf{5.45} & \cellcolor{violet!7}\textbf{95.25} & \cellcolor{violet!7}\textbf{1.24} & \cellcolor{violet!7}\textbf{4.46} & \cellcolor{violet!7}\textbf{95.95}
    \\ 
    \bottomrule
    \end{tabular}
    
    \begin{tablenotes}
      \item[\textdagger] evaluating MapAnything with poses and intrinsic.
    \end{tablenotes}

  \end{threeparttable}
\end{table*}

We evaluate our zero-shot depth estimation performance on three autonomous driving datasets not included in our training data: KITTI~\cite{geiger2012kitti}, DDAD~\cite{packnet2020ddad}, and Waymo~\cite{sun2020waymo}. 
These benchmarks represent the most common autonomous driving scenarios, including adverse weather, low-parallax motion, textureless surfaces, and dark environments.
We follow the evaluation procedure and source view selection strategy from \cite{izquierdo2025mvsanywhere}. 
We sample 16 scan lines from the KITTI and DDAD datasets and 8 from the Waymo dataset as the LiDAR prompt.
Please check our \textit{supplementary materials} for implementation details of the baseline methods.

% \sqh{
% 值得注意的是在KITTI的实验中与MVSAnywhere和
% MapAnything不同，我们使用的是offical中全部的val split，因此复现的效果会和其论文中提供的存在差异
% 除此之外 PromptDA的复现按照论文的要求对输入的稀疏prompt进行了K=4的KNN预处理操作再输入到网络中。
% }

%% threshold
\paragraph{Metrics.}
Following the established evaluation protocol of \cite{izquierdo2025mvsanywhere}, we employ three widely-used metrics to quantify the predicted $\hat{D}(t)$ and the ground truth depth $D_{{gt}}(t)$.
Mean Absolute Error (MAE) is defined per-pixel as $|\hat{D}(t) - D_{gt}(t)|$.
Absolute Relative Error (AbsRel) is defined per-pixel as $\frac{|\hat{D}(t) - D_{gt}(t)|}{D_{gt}(t)}$. Inlier Percentage ($\tau < 1.25$) is defined per-pixel as $[\max(\frac{\hat{D}(t)}{D_{gt}(t)}, \frac{D_{gt}(t)}{\hat{D}(t)}) < 1.25]$, where $[\cdot]$ is the Iverson bracket.
All three metrics are first averaged across all valid GT pixels in each test image, then across all images in the dataset.
To assess the temporal consistency of our predicted depth sequences, we further adopt the Temporal Alignment Error (TAE) metric from \cite{yang2024depthanyvideo} to quantify the reprojection error of depth maps between consecutive frames.
Please check our \textit{supplementary materials} for detailed formulations.

\paragraph{Training Details.}
Our model is trained on $4\times$A100 GPUs for 240k steps, spanning approximately 1 day.
We use a batch size of 6 at $640 \times 480$ resolution.
The matching encoder and cost volume MLPs are initiated with a learning rate of $1{e}-4$, maintained for 24k steps, then reduced to $1{e}-5$.
The Sptio-Temporal Decoder and Triple-Cue Combiner start at $5{e}-5$, linearly decay to $5{e}-8$ over the training duration.
The reference image encoder starts at $5{e}-6$ and linearly decays to $5{e}-9$.
A consistent weight decay of $1{e}-4$ is applied across the entire network.

\begin{figure*}[t]
    \begin{center}
    \includegraphics[width=\linewidth]{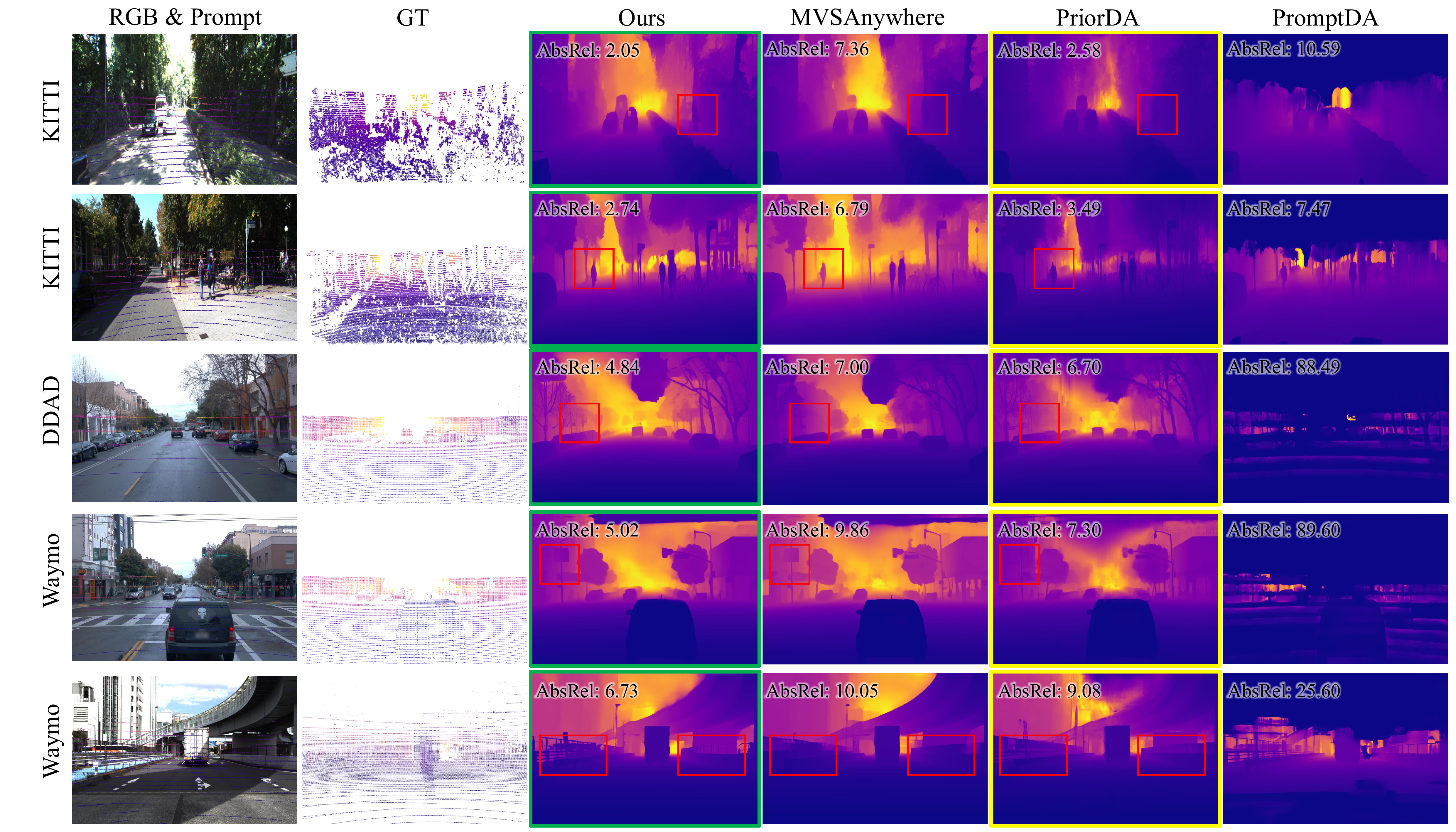}
    \end{center}
    \vspace{-0.2cm}
    \caption{
    The \textbf{qualitative results} of the estimated depth by different methods on \textit{KITTI}~\cite{geiger2012kitti}(row 1 \& 2), \textit{DDAD}~\cite{packnet2020ddad}(row 3) and \textit{Waymo}~\cite{sun2020waymo}(row 4 \& 5). 
    The \textcolor{forestgreen}{best} and \textcolor{yellowyellow}{second best} are highlighted with \textcolor{forestgreen}{green} and \textcolor{yellowyellow}{yellow} borders, respectively.
    Please check the \textcolor{red}{red} boxes in the figure for a detailed comparison.
    }
\label{fig:main}
\end{figure*}

\subsection{Experimental Results}
\paragraph{Accuracy Analysis}
As shown in \cref{tab:benchmark}, we quantitatively compare DriveMVS with a diverse set of baselines spanning various depth estimation paradigms.
Feed-forward approaches~\cite{wang2025vggt,keetha2025mapanything} enable very fast inference, yet their accuracy in autonomous driving scenarios remains suboptimal, despite improvements from supplying pose and camera intrinsics.
Monocular depth methods~\cite{wang2025moge,bochkovskii2024depthpro} benefit from large-scale training and therefore show good generalization, but as they cannot exploit multi‑view geometry or temporal cues, their accuracy lags behind methods that reason over multiple views.
Both prompt-guided~\cite{wang2025priorda} and MVS-based methods~\cite{izquierdo2025mvsanywhere} achieve strong accuracy by leveraging complementary cues, and our approach unifies these strengths by combining prompt conditioning with multi-view geometric aggregation, achieving the highest overall accuracy as shown in ~\cref{fig:main}. Feed-forward results for Waymo are omitted because processing certain sequences resulted in unavoidable GPU out‑of‑memory (OOM) errors in our test environment.

% \TODO{XX}. In contrast, monocular depth estimation methods rely heavily on their ability to infer metric-scale depth, either through strong learned priors or external metric guidance. In general, our method outperforms all the baseline methods, including both metric-prompted methods~\cite{lin2025promptda, wang2025priorda} and prior MVS-based frameworks~\cite{cao2024mvsformer++, izquierdo2025mvsanywhere}, demonstrating superior accuracy and robustness across diverse evaluation scenarios. More visualized results are presented in ~\cref{fig:main}.

\paragraph{Consistency Analysis}
% 时序对比小表
\begin{table}[t]
\caption{Comparison of temporal consistency on \textit{KITTI}~\citep{geiger2012kitti} for \textbf{video depth estimation}. All scores are given in percentage (\%). The \textbf{best} and the \underline{second best} results are highlighted in \textbf{bold} and \underline{underline}.}
\vspace{-0.2cm}
\label{tab:temporal_benchmark}
\resizebox{1\columnwidth}{!}{
\begin{tabular}{r|r|ccc}
    \toprule
    \multirow{2}{*}{\textbf{Method}} & \multirow{2}{*}{\textbf{Venue}} & \multicolumn{3}{c}{\textbf{KITTI}} \\
    & & \textbf{AbsRel~($\downarrow$)} & \textbf{$\tau$~($\uparrow$)} & \textbf{TAE~($\downarrow$)}
    \\\midrule\midrule
    % ChoronDepth~\citep{shao2025chronodepth} & CVPR'25 &  
    % \\
    % DepthCrafter~\citep{hu2025depthcrafter} & CVPR'25 &  
    % \\
    % DepthAnyVideo~\citep{yang2024depthanyvideo} & ICLR'25 &
    % \\
    VideoDA-B~\cite{chen2025videodepthanything} & CVPR'25 & 16.64 & 83.17 & 0.767
    \\
    MVSAnywhere~\cite{izquierdo2025mvsanywhere} & CVPR'25 & 10.37 & 91.05 & 0.338
    \\
    \midrule
    \cellcolor{violet!7}\textcolor{violet!66}{\textbf{Ours}} & \cellcolor{violet!7}- & \cellcolor{violet!7}\textbf{2.56} & \cellcolor{violet!7}\textbf{98.78} & \cellcolor{violet!7}\textbf{0.296}
    \\
    \bottomrule
    \end{tabular}
}
\end{table}

As shown in~\cref{tab:temporal_benchmark}, we report both depth accuracy and temporal consistency metrics, comparing our method with state-of-the-art video depth estimation~\cite{chen2025videodepthanything} and MVS-based approaches~\cite{izquierdo2025mvsanywhere}. Overall, our method outperforms all baseline methods across key metrics, demonstrating superior depth estimation accuracy and enhanced temporal coherence.

\subsection{Ablation Study}
\label{sec:ablation}

\paragraph{Ablations on Proposed Components.}
\begin{table}[t]
\caption{
Ablation studies on: \textbf{[left]} Prompt-Anchored Cost Volume (PACV), Triple-Cues Combiner (TCC), Spatio-Temporal Decoder (STD), and \textbf{[middle]} temporal loss.
}
\vspace{-0.2cm}
\label{tab:ablation_components}
\centering\scalebox{0.7}{
\begin{tabular}{c|ccc|cc|ccc}
    \toprule
    \multirow{2}{*}{\textbf{\#}} & \multirow{2}{*}{\textbf{PACV}} & \multirow{2}{*}{\textbf{TCC}} & \multirow{2}{*}{\textbf{STD}} & \multirow{2}{*}{$\mathcal{L}_{\text{s}}$} & \multirow{2}{*}{$\mathcal{L}_{\text{t}}$} & \multicolumn{3}{c}{\textbf{KITTI}} 
    \\
    & & & & & &\textbf{AbsRel}~($\downarrow$) & $\tau$~($\uparrow$) & \textbf{TAE}~($\downarrow$)
    \\\midrule\midrule
    1 &  &  &  & \checkmark & & $10.37$ & $91.05$ & $0.338$
    \\
    2 & \checkmark & & & \checkmark & & $4.86$ & $94.21$ & $0.338$
    \\
    3 & \checkmark & \checkmark & & \checkmark & & $2.76$ & $98.72$ & $0.338$
    \\
    4 & & & \checkmark & \checkmark & \checkmark & $10.18$ & $91.30$ & $0.297$ 
    \\
    5 & \checkmark & \checkmark & \checkmark & \checkmark & & $2.58$ & $98.76$ & $0.335$ 
    \\ \midrule
    \cellcolor{violet!7}\textcolor{violet!66}{\textbf{6}} & \cellcolor{violet!7}\checkmark & \cellcolor{violet!7}\checkmark & \cellcolor{violet!7}\checkmark & \cellcolor{violet!7}\checkmark & \cellcolor{violet!7}\checkmark & \cellcolor{violet!7}\textbf{2.56} & \cellcolor{violet!7}\textbf{98.78} & \cellcolor{violet!7}\textbf{0.296}
    \\
    \bottomrule
    \end{tabular}
}
\vspace{-0.0cm}
\end{table}

We validate the effectiveness of the proposed modules and training loss through extensive experiments on the KITTI dataset. As shown in ~\cref{tab:ablation_components}, our key findings include:
\begin{itemize}
    \item The \textbf{PACV} and \textbf{TCC} (Exp1\&2\&3) exploit the absolute scale cues from sparse LiDAR prompts, significantly improving metric accuracy;
    \item The \textbf{Spatio-Temporal Decoder} together with the \textbf{temporal loss} (Exp1\&4) play a crucial role in enhancing temporal smoothness, while also slightly improving per-frame depth accuracy;
    \item \textbf{Collaboration of Cost Volume and Spatio-Temporal Decoder} (Exp4\&5) leads to overall performance improvement, while the important role of temporal loss is also reflected.
\end{itemize}
The integrated system (Exp6), which combines all components, achieves the best overall performance, demonstrating the complementary contributions of each module to both depth estimation accuracy and spatio-temporal coherence.

\begin{figure}[t]
    \begin{center}
    \includegraphics[width=\linewidth]{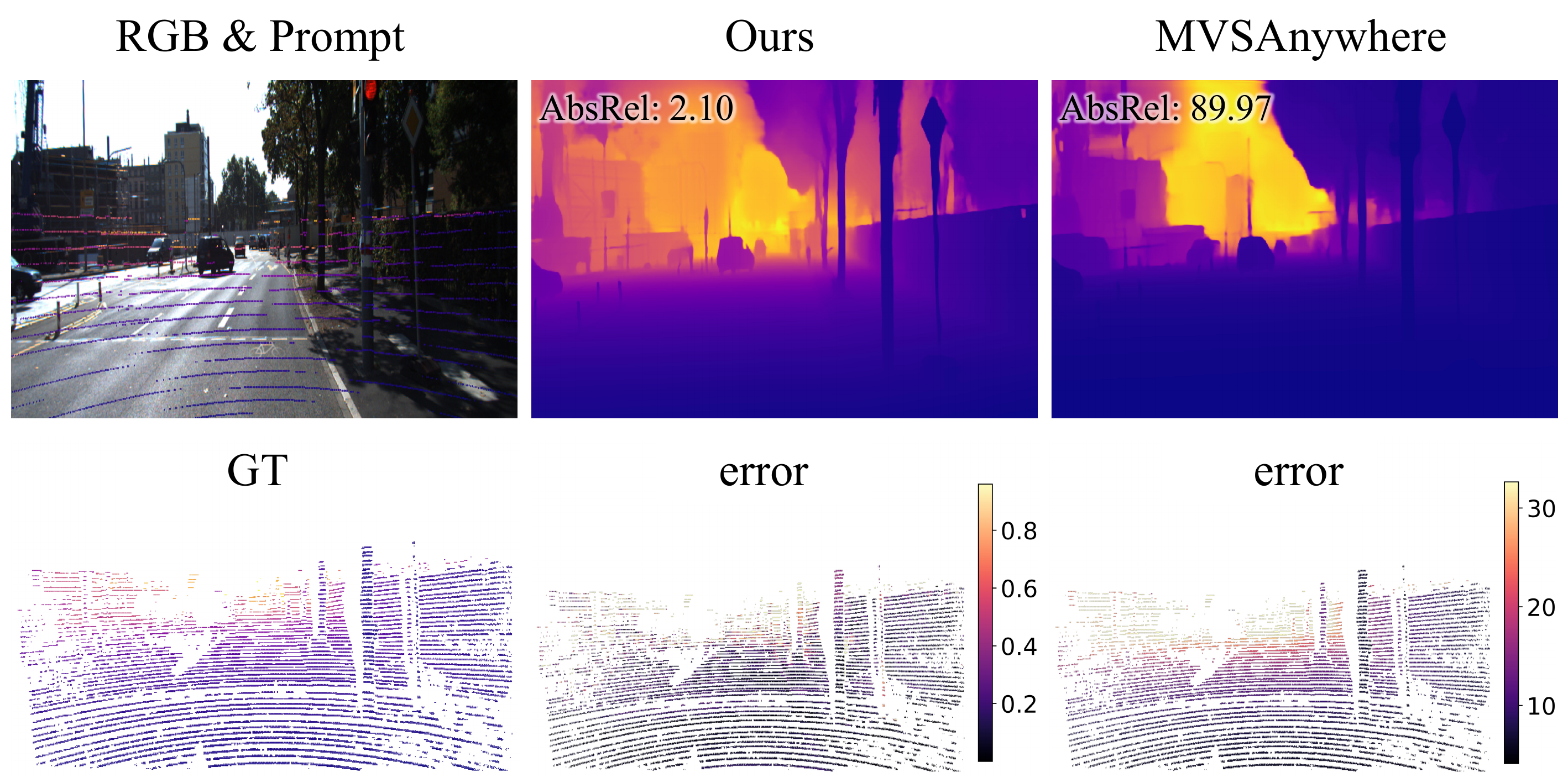}
    \end{center}
    \vspace{-0.2cm}
    \caption{Visualization of depth estimation on a static scene. The result demonstrates our robustness in challenging, low-parallax scenarios.} 
\label{fig:extreme}
\end{figure}

\paragraph{Robustness to Extreme Cases.}
% 图和表
\begin{table}[t]
    \centering
    \caption{Ablation studies on extreme cases, including adverse weather, low-lighting, and ego-static situations. The \textbf{best} and \underline{second best} scores are highlighted in \textbf{bold} and \underline{underline}.}
    \vspace{-0.1cm}
\label{tab:extreme}
\resizebox{1\columnwidth}{!}{
    \begin{tabular}{l|cc|cc|cc}
    \toprule
    \multirow{2}{*}{\textbf{Method}} & \multicolumn{2}{c}{\textbf{Rainy}} \vline & \multicolumn{2}{c}{\textbf{Dark}} \vline & \multicolumn{2}{c}{\textbf{Static}}
    \\
    & \textbf{AbsRel}($\downarrow$) & \textbf{$\tau$}($\uparrow$) & \textbf{AbsRel}($\downarrow$) & \textbf{$\tau$}($\uparrow$) & \textbf{AbsRel}($\downarrow$) & \textbf{$\tau$}($\uparrow$)
    \\\midrule\midrule
    DepthPro~\citep{bochkovskii2024depthpro} & $34.96$ & $19.30$ & $43.63$ & $13.43$ & $34.03$ & $11.23$ \\
    PromptDA~\citep{lin2025promptda} & $86.15$ & $6.44$ & $81.60$ & $10.06$ & $88.15$ & $2.46$ \\
    PriorDA~\citep{wang2025priorda} & \underline{$11.92$} & \underline{$86.56$} & \underline{$11.25$} & \underline{$83.59$} & \underline{$6.78$} & \underline{$93.30$} \\
    MVSAnywhere~\citep{izquierdo2025mvsanywhere} & $18.48$ & $82.85$ & $16.97$ & $79.74$ & $55.56$ & $43.23$ \\
    \midrule
    \cellcolor{violet!7}\textcolor{violet!66}{\textbf{Ours}} & \cellcolor{violet!7}\textbf{7.21} & \cellcolor{violet!7}\textbf{94.02} & \cellcolor{violet!7}\textbf{4.97} & \cellcolor{violet!7}\textbf{94.84} & \cellcolor{violet!7}\textbf{4.93} & \cellcolor{violet!7}\textbf{95.56}
    \\    
    \bottomrule
\end{tabular}
}
\end{table}
As shown in \cref{tab:extreme}, we conduct extensive experiments on extreme cases in autonomous driving, including rainy weather, dark environments, and ego-static situations, to demonstrate the robustness of DriveMVS. In these low-parallax and texture-less scenes, previous methods suffer from varying degrees of degradation due to their lack of absolute scale, multiple perspectives, or temporal information. \cref{fig:extreme} shows an example of an ego-static scene, where MVSAnywhere~\cite{izquierdo2025mvsanywhere} performs inferior, and our method maintains robustness to these corner cases. Please check our \textit{supplementary materials} for detailed settings.

\begin{figure}[t]
    \begin{center}
    \includegraphics[width=\linewidth]{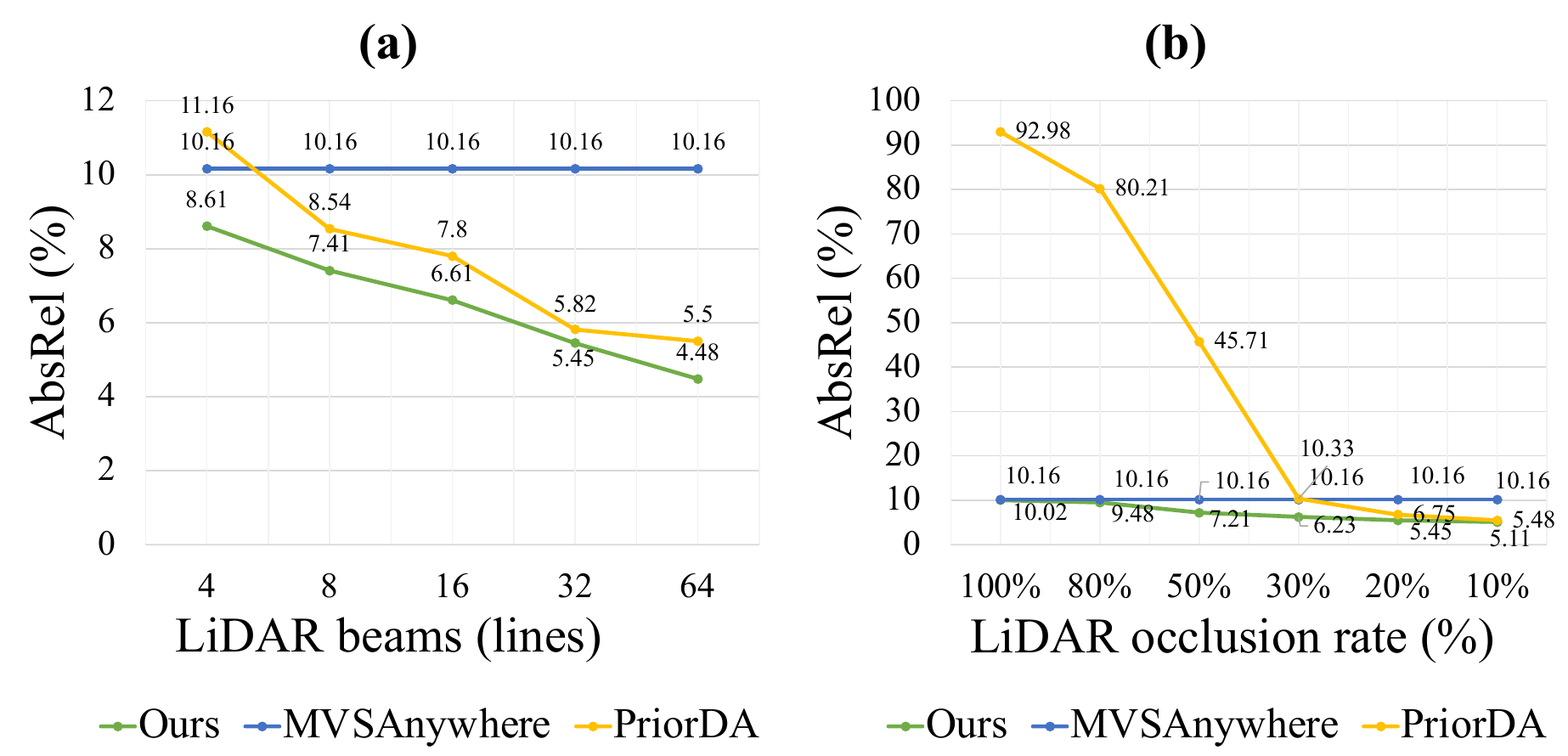}
    \end{center}
    \vspace{-0.2cm}
    \caption{Ablations on (a) different LiDAR laser beams and (b) different LiDAR occlusion rate (from bottom). \textbf{Lower AbsRel refers to a better result.}} 
\label{fig:absence}
\end{figure}

\paragraph{Robustness to Prompt Density and Occlusion.}
To further illustrate the effectiveness of our method in an actual autonomous driving scene, we degrade the LiDAR prompt by varying degrees to simulate the loss of the original prompt, and conduct ablation experiments across different datasets. As shown in \cref{fig:absence}(a), we sparsify the LiDAR laser beams from 64 lines to 4 lines. Results indicate that our method consistently performs better. We further apply a bottom-up occlusion to the LiDAR scan lines in each frame to mimic near-field blind spots caused by ego-vehicle self-occlusion. \cref{fig:absence}(b) shows that across a range of occlusion ratios, our method remains robust and maintains a clear advantage.

\begin{figure}[t]
    \begin{center}
    \includegraphics[width=\linewidth]{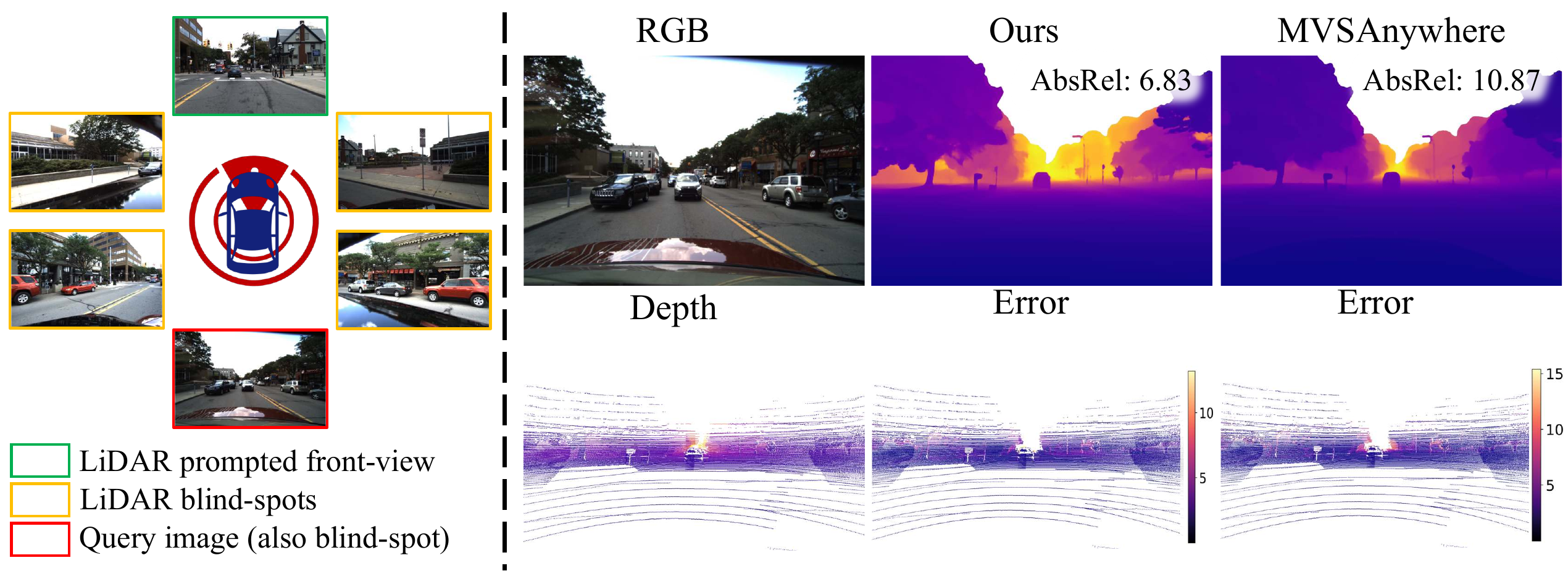}
    \end{center}
    \vspace{-0.2cm}
    \caption{Ablation study on Lidar prompt absence. We estimate the depth of the back-view image while only the front-view image is provided by LiDAR. Our method remains an accurate metric under such a configuration.} 
\label{fig:multi-view}
\end{figure}

\paragraph{Robustness to Prompt Absence.}
We also simulated the blind-spot problem caused by the single forward-looking LiDAR configuration on the DDAD dataset. As shown in \cref{fig:multi-view}, for a sensor system composed of 6 cameras and a single forward-looking LiDAR, the LiDAR prompt in the front main view is the only perspective that is not a potential blind spot without a prompt. We compare our method with MVSAnywhere~\cite{izquierdo2025mvsanywhere}, and the results show that our method achieves better depth accuracy and far exceeds the baseline when combining an absolute metric prompt with multi-view clues. Please check our \textit{supplementary materials} for more experiment results.

% 线束，视角，盲区
% \input{tables/num_lines}
% KITTI 线束的影响折线图（PromptDA， PriorDA，Ours，MVSAnywhere），

% Benchmark
% 有无prompt的影响
% 静态场景的影响
% kitti-lines的影响
% 参考图数量的影响
% 时序数量的影响
% 作为补全算法的后续修正
% 训练过程中融入真实场景以及非zero-short的训练
% 不同prompt类型的影响 (VOID)
% 训练的数据量对KITTI-DDAD的结果影响
% 时序Window的影响

\section{Conclusion}
\label{sec:conclusion}
In this work, we presented DriveMVS, a novel multi-view stereo framework for autonomous driving. DriveMVS effectively unifies metric accuracy, temporal consistency, robustness, and cross-domain generalization. Our core innovation lies in a dual-pathway LiDAR prompt integration (via a prompt-anchored cost volume and a triple-cue combiner) combined with explicit temporal context modeling. Experiments confirm that DriveMVS achieves state-of-the-art accuracy and stability, with strong generalization, highlighting its potential for scalable 3D perception.

\paragraph{Limitations.} DriveMVS's inference time is currently higher than monocular methods due to its multi-view and temporal dependencies. Future work will target optimizing the computational efficiency of the MVS backbone.

% In this work, we present MVS-Pro, a novel multi-view stereo framework tailored for autonomous driving under minimalist LiDAR configurations. By unifying monocular priors, multi-view geometry, sparse metric prompts, and temporal context within a single architecture, MVS-Pro effectively reconciles the long-standing trade-off between metric accuracy, temporal consistency, robustness, and zero-shot cross-domain generalization. Our key insight is to treat sparse LiDAR observations as dual-path anchors—providing hard geometric constraints through a prompt-anchored cost volume and soft feature-level guidance via a triple-cue combiner—while explicitly modeling temporal context to ensure stable depth propagation across frames, even in the absence of strong parallax or optical flow. Extensive experiments on autonomous driving benchmarks show that MVS-Pro achieves state-of-the-art performance in both per-frame accuracy and sequence-level stability, with strong generalization across domains and resilience to sensor sparsity, highlighting its potential for scalable and reliable 3D perception.

% \paragraph{Limitations.}
% Due to the reliance of our method on multi perspective or temporal information to construct the cost volume during inference, our method is slightly inferior to traditional monocular depth estimation in terms of inference time. Future work may include optimizing the computational cost of the MVS backbone.

\section*{Acknowledgments}

We thank reviewers for their constructive comments. This research was supported by MSSJH20230070.

{
    \small
    \bibliographystyle{ieeenat_fullname}
    \bibliography{main}
}

\clearpage
\setcounter{page}{1}
\maketitlesupplementary
\appendix

In this document, we further provide the following materials to support the statements and conclusions drawn in the main body of this paper.
Please find more video results in our \textit{supplementary video}.
\begin{itemize}
    \item \cref{sec:model_and_training}: Model and Trining Details;
    \item \cref{sec:implementation_details}: Implementation Details;
    \item \cref{sec:more_discussions}: More Discussions of Experiments.
\end{itemize}

\section{Model and Training Details}
\label{sec:model_and_training}

\subsection{Model Architecture}

\paragraph{Sparsity-aware Prompt Encoder.}

To effectively leverage sparse LiDAR points as high-fidelity metric prompts, we introduce a Sparsity-aware Prompt Encoder. 
A standard CNN encoder with strided convolutions or average pooling is ill-suited for this task, as it tends to dilute valid sparse signals with zeros from empty regions, leading to signal loss and feature blurring. 
Therefore, our prompt encoder is specifically designed to preserve the integrity of sparse signals during the downsampling process while aligning them with the network's token space.

Specifically, given a raw sparse metric prompt with corresponding valid mask $\mathbf{P} \in \mathbb{R}^{2 \times H \times W}$, we first transform it into the logit space to match the network's prediction target. 
To incorporate spatial awareness, we concatenate normalized coordinate grids $(\mathbf{Y}, \mathbf{X})$ to the input, resulting in an augmented input feature $\mathbf{X}_{in} \in \mathbb{R}^{(2+2) \times H \times W}$.
Then $\mathbf{X}_{in}$ is processed by a stem block followed by three downsampling stages. 
To prevent signal degradation, we replace standard pooling layers with a custom Masked Max-Pooling operation.
Let $\mathbf{F}_l$ be the feature map and $\mathbf{M}_l$ be the binary validity mask at stage $l$. 
The downsampling process is defined as:
\begin{equation}
    \mathbf{F}_{l+1}, \mathbf{M}_{l+1} = \text{MaskedMaxPool}(\text{ConvBlock}(\mathbf{F}_l), \mathbf{M}_l),
\end{equation}
where $\text{MaskedMaxPool}$ performs max pooling only on valid pixels (where $\mathbf{M}_l=1$). 
Specifically, invalid positions in the pooling window are masked with $-\infty$ before the max operation, ensuring that even a single valid signal within a window is propagated to the next resolution.
After four processing stages, the feature map is downsampled to $1/16$ of its original resolution. 
A $1\times1$ convolution projects the features to the embedding dimension $D$. 
We flatten the spatial dimensions to obtain a sequence of tokens $\mathbf{F}_{metric} \in \mathbb{R}^{ N\times D}$. 
To retain spatial context after flattening, we add sinusoidal positional embeddings of 2D absolute positions to the tokens.
Finally, to ensure the subsequent Transformer layers do not attend to empty regions, we generate a token-level attention mask $\mathbf{M}_{attn}$ derived from the final downsampled validity mask. 
The output features $\mathbf{F}_{metric}$ are masked, setting invalid regions to zero, forcing the network to rely solely on valid metric cues.

\begin{figure}[t]
    \begin{center}
    \includegraphics[width=\linewidth]{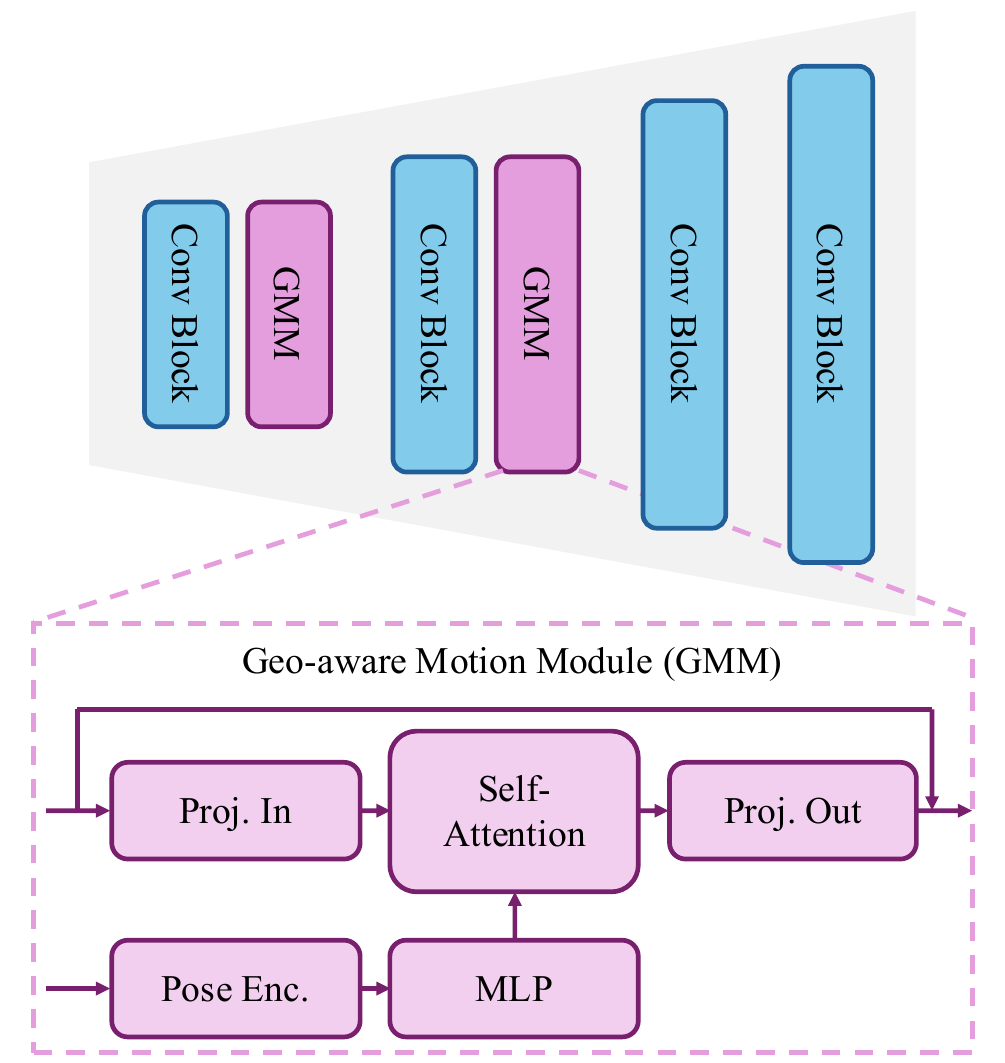}
    \end{center}
    \vspace{-0.2cm}
    \caption{\textbf{Illustrations of spatio-temporal decoder.}
    Please refer to~\cref{sec:model_and_training} for more details.
    }
\label{fig:std}
\end{figure}

\paragraph{Spatio-Temporal Decoder.}
While the Triple-Cues Combiner (\cref{sec:triple_cues_combiner}) effectively fuses multi-modal cues within a single frame, independent decoding fails to guarantee temporal consistency across the video sequence.
Standard video transformers rely on blind learnable positional encodings, ignoring the explicit 3D geometric relationships provided by camera poses.
To address this, we propose a Spatio-Temporal Decoder.
As illustrated in~\cref{fig:std}, it builds upon the Dense Prediction Transformer (DPT) architecture~\cite{ranftl2021dpt}, with Geometry-Aware Motion Modules (GMMs) that extend it from static image processing to geometry-aware video sequence modeling. 
Instead of asking the transformer to hallucinate 3D relationships from 2D sequences, we explicitly embed each pixel's 3D geometric state into the decoder's features.
For each pixel $\mathbf{p}=(u,v)$ in frame $t$, we compute its camera ray origin $\mathbf{o}_t \in \mathbb{R}^3$ and normalized direction $\mathbf{d}_t \in \mathbb{R}^3$ in the world coordinate system
\begin{equation}
    \mathbf{d}_t = \mathbf{R}_t \mathbf{K}^{-1} \mathbf{p}, \quad \mathbf{o}_t = \mathbf{C}_t,
\end{equation}
where $\mathbf{R}_t, \mathbf{C}_t$ are the camera rotation and center derived from the pose.
To allow the network to learn high-frequency geometric details (e.g., small parallax changes), we map the 6D ray coordinate $[\mathbf{o}_t, \mathbf{d}_t]$ to a high-dimensional feature space using Fourier features:
\begin{equation}
    \gamma(\mathbf{v}) = [\dots, \sin(2^k \pi \mathbf{v}), \cos(2^k \pi \mathbf{v}), \dots].
\end{equation}
This embedding is then projected to the feature dimension $C$ via a shallow MLP:
\begin{equation}
    E_{geo}(t, u, v) = \text{MLP}(\gamma([\mathbf{o}_t, \mathbf{d}_t])).
\end{equation}
We inject this geometric embedding directly into the decoder features before the temporal attention layers
\begin{equation}
\begin{aligned}
    &\hat{F} = F + E_{geo}.
\end{aligned}
\end{equation}
When the transformer computes attention between frame $t$ and $t'$, it can now explicitly compare their ray embeddings. 
Tokens with intersecting or converging rays (indicating the same 3D surface point) will naturally have higher affinity, thereby enforcing multi-view 3D consistency in a geometrically grounded manner.

\subsection{Training Details}
For network training, we utilized four synthetic datasets with precise depth annotations: TartanAir~\cite{wang2020tartanair}, VKITTI~\cite{cabon2020vkitti2}, TartanGround~\cite{patel2025tartanground}, and MVS-Synth~\cite{huang2018deepmvs}, totaling $2.03$ million samples. 
It is worth noting that for the VKITTI dataset, we incorporated both left- and right-camera views to accommodate multi-view inputs, effectively doubling the original VKITTI data.
During training, sparse depth maps are synthetically generated by sub-sampling the dense ground truth. 
We simulate random 8–64 line LiDAR patterns with variations in angle and shift.
To mimic the inherent measurement fluctuations found in hardware sensors, we simulate sensor precision errors by applying Gaussian noise to the radial distance of each point along its projection ray.
To simulate signal loss due to reflective surfaces or distance attenuation, we randomly discard a percentage of the input points, encouraging the model to be robust against varying point densities and missing data.

To enable parallel imitation of interaction logic for multiple targets, we apply specific rules when selecting targets in the data pipeline. 
There is a 0.5 probability that the target exists in both modalities, a 0.25 probability that only the current frame has a prompt, and a 0.25 probability that only the source frames have prompts.

DriveMVS is trained for 240k steps at an image resolution of 640 × 480, with prompts resized to match. 
The network is optimized using AdamW~\cite{loshchilov2017adamw} with a OneCycleLR~\cite{smith2019onecyclelr} scheduler. 
Data augmentation includes horizontal image flipping (probability of $0.5$), along with the remaining hyperparameters, following the configuration in MVSAnywhere~\cite{izquierdo2025mvsanywhere}.

\subsection{Loss Functions}
We train DriveMVS using a compound objective function that enforces both spatial geometric fidelity and temporal coherence.
\paragraph{Spatial Geometry Losses.}
To ensure precise per-frame reconstruction, we adopt a multi-scale supervision strategy inspired by~\cite{izquierdo2025mvsanywhere}.
First, we apply a Log-Depth $\mathcal{L}_1$ Loss to compute the absolute error between the logarithm of the ground truth $D_{gt}$ and the predicted depth $\hat{D}$ across four decoder scales $s$:
\begin{equation}
\mathcal{L}_{depth} = \frac{1}{HW} \sum_{s=1}^{4}\sum_{i,j} \frac{1}{s^2} \left| \uparrow_{gt} \log \hat{D}{r}^{i,j} - \log D^{i,j} \right|,
\end{equation}
where $\uparrow_{gt}$ denotes nearest-neighbor upsampling to the ground truth resolution.
Second, to encourage sharp discontinuities and preserve high-frequency details, we utilize a Gradient Loss~\citep{izquierdo2025mvsanywhere, sayed2022simplerecon}. 
Unlike standard approaches, we compute spatial gradients ($\nabla$) in the inverse depth space to prevent the loss from being dominated by distant regions:
\begin{equation}
\mathcal{L}_{grad} = \frac{1}{HW} \sum_{s=1}^{4}\sum_{i,j} \left| \nabla\downarrow_s \frac{1}{\hat{D}_r^{i,j} } - \nabla\downarrow_s \frac{1}{D^{i,j} } \right|.
\end{equation}
Third, we enforce local surface consistency via a Surface Normal Loss~\citep{izquierdo2025mvsanywhere}. 
We minimize the cosine distance between the ground truth normals $N$ and the predicted normals $\hat{N}$, which are analytically derived from $\hat{D}$ using camera intrinsics:
\begin{equation}
\mathcal{L}_{normals} = \frac{1}{2HW} \sum_{i,j} \left( 1 - \hat{N}^{i,j} \cdot N^{i,j} \right).
\end{equation}

\paragraph{Temporal Consistency Loss.}
In order to ensure smooth depth transitions across frames, we incorporate the Temporal Gradient Matching (TGM) Loss~\citep{chen2025videodepthanything}. 
This objective enforces temporal consistency between the predicted depth sequence and the ground truth, without requiring optical flow estimation. 
The temporal loss is defined as:
\begin{equation}
\begin{aligned}    
\mathcal{L}_{temporal} = & \frac{1}{T-1} \sum_{t=1}^{T-1} \sum_{i,j} \\
& \left\| \left| \hat{D}_{t+1}^{i,j} - \hat{D}_{t}^{i,j} \right| - \left| D_{t+1}^{i,j} - D_{t}^{i,j} \right| \right\|_1.
\end{aligned}
\end{equation}
Specifically, we apply a masking threshold $\tau_{temp}=0.05$ to the ground truth temporal change ($|D_{t+1} - D_t| < \tau_{temp}$) to filter out regions with occlusions or dynamic objects, ensuring the loss focuses on static geometric consistency.

\paragraph{Total Loss.}
The final training objective is a weighted sum of the spatial and temporal components, as described in \cref{eq:total-loss} of the main paper.

\section{Implementation Details}
\label{sec:implementation_details}

\subsection{Data Preprocessing}
\label{sec:data-preprocessing}
To validate the feasibility of the proposed solution, we utilize front-view camera images and LiDAR data from the Waymo~\cite{sun2020waymo}, KITTI~\cite{geiger2012kitti}, and DDAD~\cite{packnet2020ddad} datasets. 
This setup is designed to be extendable to multi-camera and LiDAR configurations in future work.
It is worth noting that while KITTI provides both accumulated dense ground truth (from multi-frame stitching) and single-frame raw LiDAR, DDAD and Waymo only provide per-frame 128-beam and 64-beam LiDAR data, respectively. 
Given the susceptibility of multi-frame accumulation to artifacts from dynamic objects, we consistently use single-frame LiDAR as the ground truth across all datasets.
For prompt generation, we back-project the single-frame LiDAR data into 3D space to calculate beam inclination angles. 
Based on these angles, we downsample the data to 16, 8, and 8 beams, respectively, to serve as the sparse LiDAR prompts.

\subsection{Metrics}
\begin{table}[htbp]
\centering
\caption{
\textbf{Depth metric definitions.}
$D_{gt}$ and $\hat{D}$ are the ground-truth and predicted depth, respectively.
}
\scalebox{1.0}{%
\begin{tabular}{ll}
\toprule
\textbf{Metric} & \textbf{Definition} \\
\midrule\midrule		
MAE & $\frac{1}{N}\sum_{i=1}^N|D_{gt}-\hat{D}|$ \\
AbsRel & $\frac{1}{N}\sum_{i=1}^N|D_{gt}-\hat{D}|/D_{gt}$\\
$\tau$ & $\frac{1}{N}\sum_{i=1}^N\left(\max\left(\frac{\hat{D}}{D_{gt}},\frac{D_{gt}}{\hat{D}}\right) < 1.25\right)$ \\
\bottomrule
\end{tabular}
}

\label{tab:depth_metric}
\end{table}

For depth metrics, we report MAE, AbsRel, and $\tau$. Their definitions can be found in~\cref{tab:depth_metric}.
For temporal consistency metrics, we report TAE defined as:
\begin{equation}
\begin{aligned}
\text{TAE} = \frac{1}{2N} &\sum_{k=1}^N(\text{AbsRel}(f(\hat{x}_d, p^k), \hat{x}_d^{k+1}) + \\ 
&\text{AbsRel}(f(\hat{x}_d^{k+1},p_{-}^{k+1}),\hat{x}_d^k)),
\end{aligned}
\end{equation}
where, $f$ represents the projection function that maps the depth $\hat{x}_d^k$ from the $k$-th frame to the $(k+1)$-th frame using the transform matrix $p^k$.
$p_{-}^{k+1}$ is the inverse matrix for inverse projection.
$N$ denotes the number of frames.
% For reconstruction metrics, we report Acc, Comp, Prec, Recall, and F-score.
% Their definitions can be found in~\cref{tab:recon_metric}.
% We use a voxel size of 0.04m for TSDF reconstruction.

\subsection{Baseline Implementation}
To ensure fair comparison, we evaluate all baselines on same test sequences according to their official repository and open source model, including VGGT~\cite{wang2025vggt}\footnote{https://github.com/facebookresearch/vggt}, MapAnything~\cite{keetha2025mapanything}\footnote{https://github.com/facebookresearch/map-anything}, MoGe-2~\cite{wang2025moge}\footnote{https://github.com/microsoft/MoGe}, DepthPro~\cite{bochkovskii2024depthpro}\footnote{https://github.com/apple/ml-depth-pro}, PromptDA~\cite{lin2025promptda}\footnote{https://github.com/DepthAnything/PromptDA}, PriorDA~\cite{wang2025priorda}\footnote{https://github.com/SpatialVision/Prior-Depth-Anything}, MVSFormer++~\cite{cao2024mvsformer++}\footnote{https://github.com/maybeLx/MVSFormerPlusPlus}, MVSAnywhere~\cite{izquierdo2025mvsanywhere}\footnote{https://github.com/nianticlabs/mvsanywhere} and VideoDepthAnything~\cite{chen2025videodepthanything}\footnote{https://github.com/DepthAnything/Video-Depth-Anything}. Specifically, we implement MapAnything under two different settings: one without camera poses and intrinsics as pure feed-forward setting, and one with camera poses and intrinsics as MVS-like setting. For VideoDepthAnything, we select the base model for evaluation. Notice that the original PromptDA and PriorDA require dense depth as a metric prompt. To fit our autonomous driving settings, we use the same LiDAR prompt as our method (described in \cref{sec:data-preprocessing}).

\section{More Discussions of Experiments}
\label{sec:more_discussions}

\subsection{Analysis of Baseline Performance}
To ensure a fair comparison, we reproduced all baseline methods on a single A100 GPU, strictly adhering to their officially provided pretrained weights and configurations.
It is crucial to note that, unlike MVSAnywhere~\cite{izquierdo2025mvsanywhere} and MapAnything~\cite{keetha2025mapanything}, for our KITTI experiments, we use the entire official validation split for evaluation.
Consequently, our reported performance metrics for these baselines may differ from those presented in their respective publications, which often employ a subset or an alternative split.
We present a quantitative comparison against state-of-the-art baselines on three challenging autonomous driving datasets: \textit{KITTI}, \textit{DDAD}, and \textit{Waymo}. 
As summarized in \cref{tab:benchmark}, our method consistently outperforms all competing approaches across all metrics and datasets, establishing a new state of the art for robust metric depth estimation.

\paragraph{Comparison with Feed-forward Reconstruction.}
Feed-forward approaches such as VGGT~\cite{wang2025vggt} and MapAnything~\cite{keetha2025mapanything} prioritize speed but compromise on reconstruction accuracy. 
The substantial error observed in VGGT (13.19m MAE on KITTI) can be attributed to the difficulty of direct geometry regression, which overlooks the explicit constraints provided by camera intrinsics and poses. 
Even when MapAnything is augmented with ground-truth poses (MapAnything$^\dagger$), improving its MAE to 1.27m, it remains inferior to our method. 
This comparison validates our design choice to leverage MVS formulations, which are essential for delivering stable, metric-scale geometry in autonomous driving scenarios.

\paragraph{Comparison with Monocular Methods.}
Monocular methods without prompts, such as MoGe-2~\cite{wang2025moge} and DepthPro~\cite{bochkovskii2024depthpro}, aim to directly recover metric depth and demonstrate impressive generalization.
However, they suffer from inherent scale ambiguity in complex, large-scale autonomous driving environments, resulting in poor metric accuracy (\eg, DepthPro achieves only 80.71\% inlier ratio on KITTI, compared to our 98.78\%).
When integrating sparse LiDAR prompts, PromptDA~\cite{lin2025promptda} still performs inferiorly due to its strong dependency on dense depth prompts.
We attribute this to its architectural design, which targets indoor scenarios and operates more like depth super-resolution, heavily relying on dense (albeit coarse) depth observations rather than the sparse inputs typical of autonomous driving. 
It is worth noting that, strictly following the official implementation, we applied KNN ($k=4$) densification to the sparse prompts for PromptDA to ensure a fair comparison, yet it failed to yield satisfactory results.
In contrast, while PriorDA~\cite{wang2025priorda} serves as a competitive baseline, our method outperforms it by a substantial margin (\eg, reducing MAE by $\sim$20\% on KITTI and $\sim$31\% on Waymo). 
This significant improvement validates the indispensability of the explicit geometric cues provided by an MVS backbone for accurate depth recovery.

\paragraph{Comparison with MVS Baselines.}
Traditional MVS-based methods like MVSFormer++~\cite{cao2024mvsformer++} and the recent generalist MVSAnywhere~\cite{izquierdo2025mvsanywhere} are susceptible to scale collapse in low-parallax scenarios (\eg, highway scenes or ego-static scenes) or textureless regions.
On Waymo, MVSAnywhere achieves a $\tau$ of 89.80\%, whereas our method reaches 95.95\%. 
This gap confirms the effectiveness of our dual-branch strategy: our method leverages MVS for geometry consistency while incorporating the TCC module to resolve ambiguities using structural and metric prompts when MVS cues are unreliable.
% As a multi-view stereo (MVS) method, MVSFormer lags behind monocular depth estimation approaches on several metrics. We attribute this primarily to its limited generalization ability and its failure to account for dynamic objects.

\paragraph{Cross-Domain Generalization.}
Crucially, our method maintains top-tier performance across diverse sensor configurations, from the 64-beam LiDAR in KITTI to the sparser setups in DDAD and Waymo (downsampled for evaluation). 
The consistent superiority across datasets validates that our model successfully generalizes, robustly fusing heterogeneous cues regardless of the specific domain shift.

To summarize, the results in \cref{tab:benchmark} validate our core hypothesis: neither MVS geometry nor sparse metric prompts are sufficient in isolation. 
MVS provides dense constraints but lacks robustness in degenerate motions; sparse prompts provide absolute scale but lack spatial density. 
By unifying these via our \textit{Prompt-Anchored Cost Volume} and \textit{Triple-Cue Combiner}, our framework effectively produces depth maps that are both metrically accurate (low MAE and AbsRel) and structurally precise (high $\tau$). 

\begin{table}[t]
    \centering
    \caption{Extreme Cases IDs
    }
    \vspace{-0.1cm}
\label{tab:extreme_ids}
\scalebox{0.8}{
    \begin{tabular}{l|cc}
    \toprule
    \multirow{2}{*}{\textbf{Case}} & \multirow{2}{*}{\textbf{Scene ID}} & \textbf{\# Total} \\
    & &\textbf{Images} \\
    \midrule\midrule
    \multirow{3}{*}{Static} & 2011\_09\_29\_drive\_0026\_sync & 148\\
    & 13941626351027979229 & 199 \\
    & 14127943473592757944 & 197 \\
    \midrule
    \multirow{4}{*}{Rainy} & 10448102132863604198 & 183 \\
    & 11356601648124485814 & 199 \\
    & 13184115878756336167 & 199 \\
    & 13415985003725220451 & 199 \\
    \midrule
    \multirow{4}{*}{Dark} & 11901761444769610243 & 196 \\
    & 13184115878756336167 & 199 \\
    & 13694146168933185611 & 193 \\
    & 14107757919671295130 & 194 \\    
    \bottomrule
\end{tabular}
}
\end{table}

\begin{figure*}[t]
    \begin{center}
    \includegraphics[width=\linewidth]{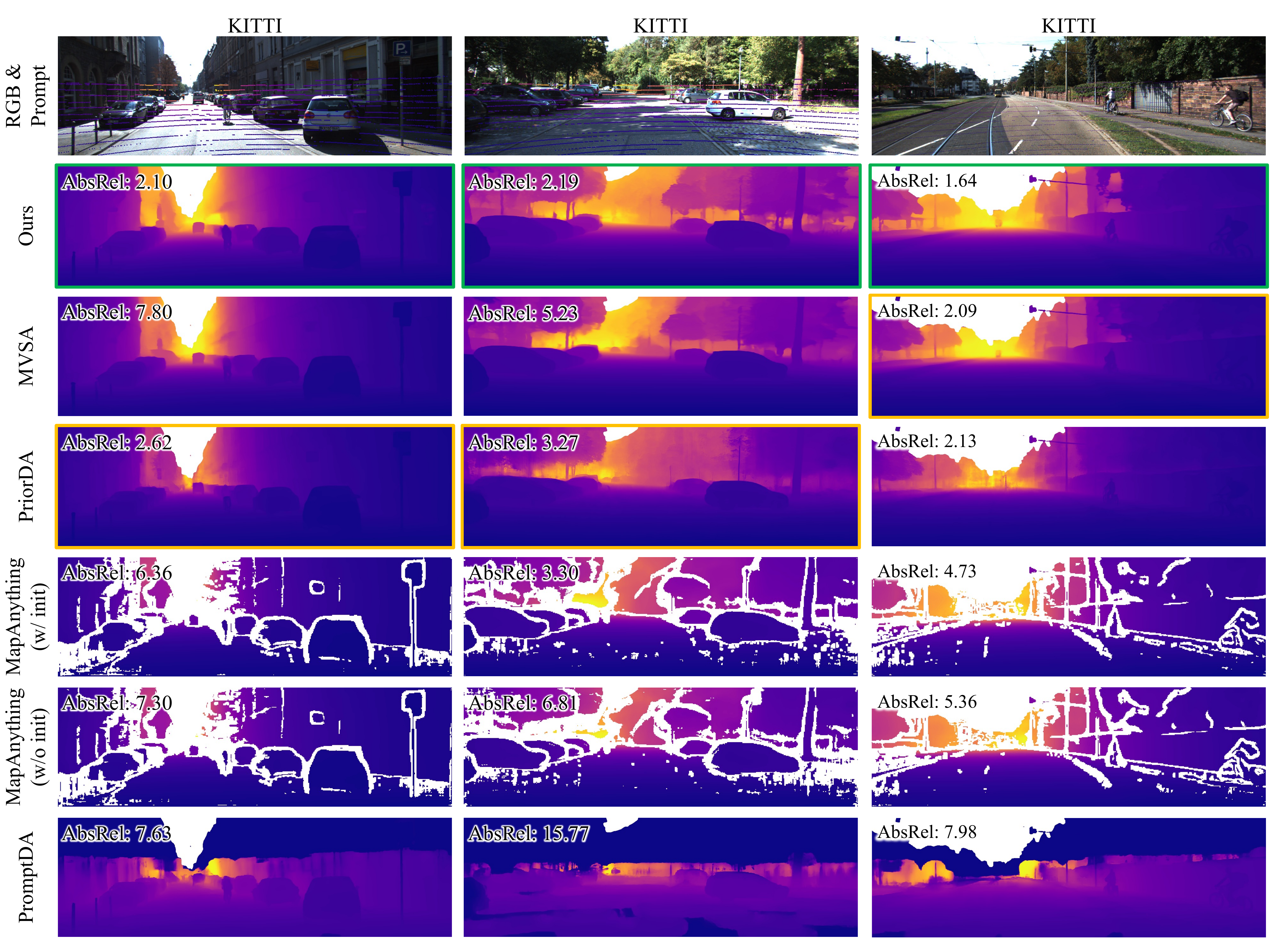}
    \end{center}
    \vspace{-0.2cm}
    \caption{
    \textbf{Qualitative comparison of depth prediction results on \textit{KITTI}~\cite{geiger2012kitti}.}
    Rows show different methods: our DriveMVS model, MVSA~\cite{izquierdo2025mvsanywhere}, PriorDA~\cite{wang2025priorda}, MapAnything~\cite{keetha2025mapanything}, and PromptDA~\cite{lin2025promptda}, along with RGB and prompt inputs ($\boldsymbol I_r$, $\boldsymbol P_r$).
    \textbf{[left]}: \textit{2011\_09\_26\_drive\_0095\_sync}.
    \textbf{[Middle]}: \textit{2011\_09\_26\_drive\_0023\_sync}.
    \textbf{[Right]}: \textit{2011\_09\_26\_drive\_0002\_sync}.
    The \textcolor{forestgreen}{best} and \textcolor{yellowyellow}{second best} are highlighted with \textcolor{forestgreen}{green} and \textcolor{yellowyellow}{yellow} borders, respectively.
    }
\label{fig:sup_depth_1}
\end{figure*}

\begin{figure*}[t]
    \begin{center}
    \includegraphics[width=\linewidth]{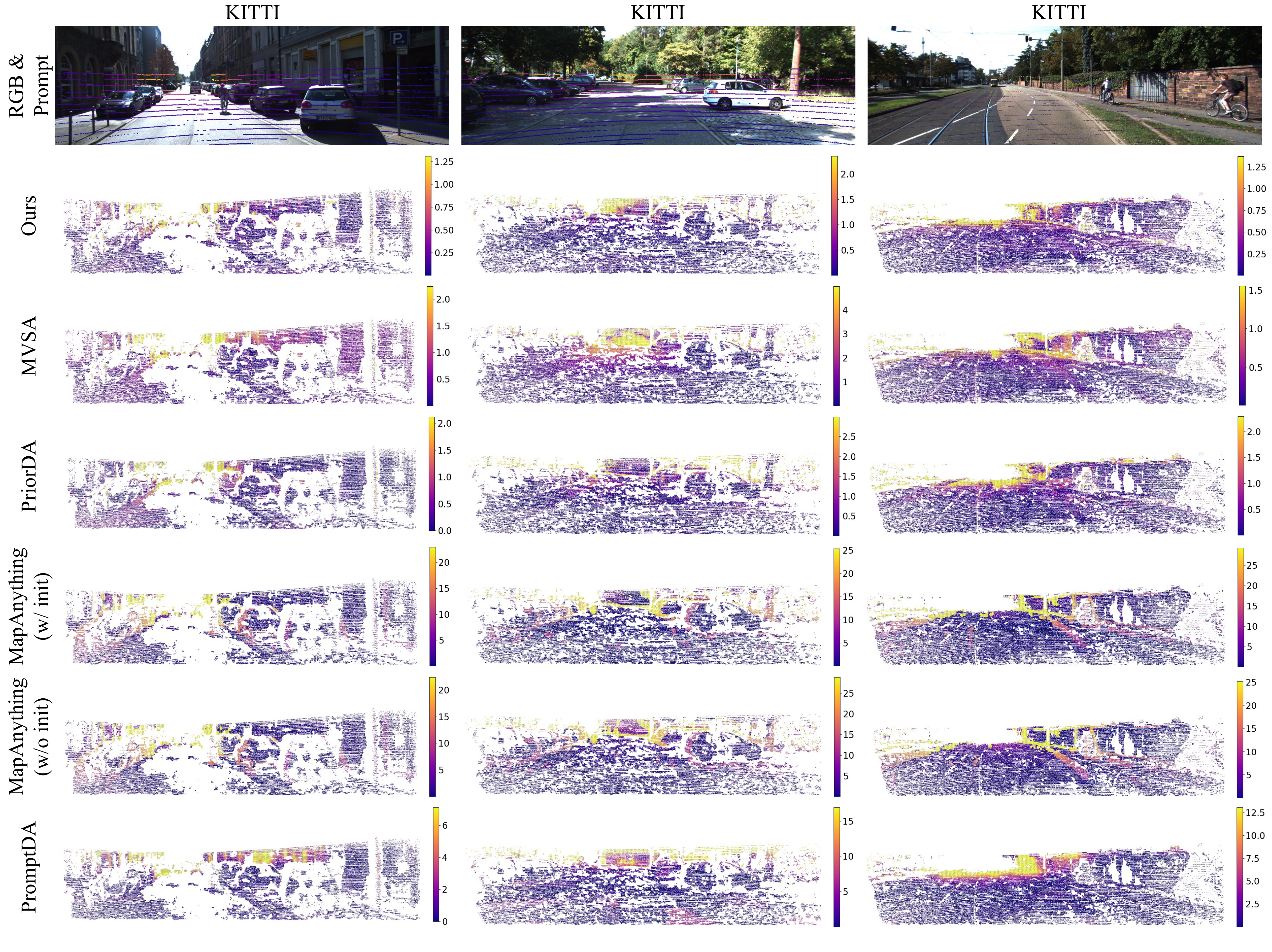}
    \end{center}
    \vspace{-0.2cm}
    \caption{
    \textbf{Qualitative comparison of depth prediction errors on \textit{KITTI}~\cite{geiger2012kitti}}, corresponding to \cref{fig:sup_depth_1}.
    Rows show different methods: our DriveMVS model, MVSA~\cite{izquierdo2025mvsanywhere}, PriorDA~\cite{wang2025priorda}, MapAnything~\cite{keetha2025mapanything}, and PromptDA~\cite{lin2025promptda}, along with RGB inputs ($\boldsymbol I_r$) and prompt ($\boldsymbol P_r$).
    }
\label{fig:sup_error_1}
\end{figure*}

\begin{figure*}[t]
    \begin{center}
    \includegraphics[width=\linewidth]{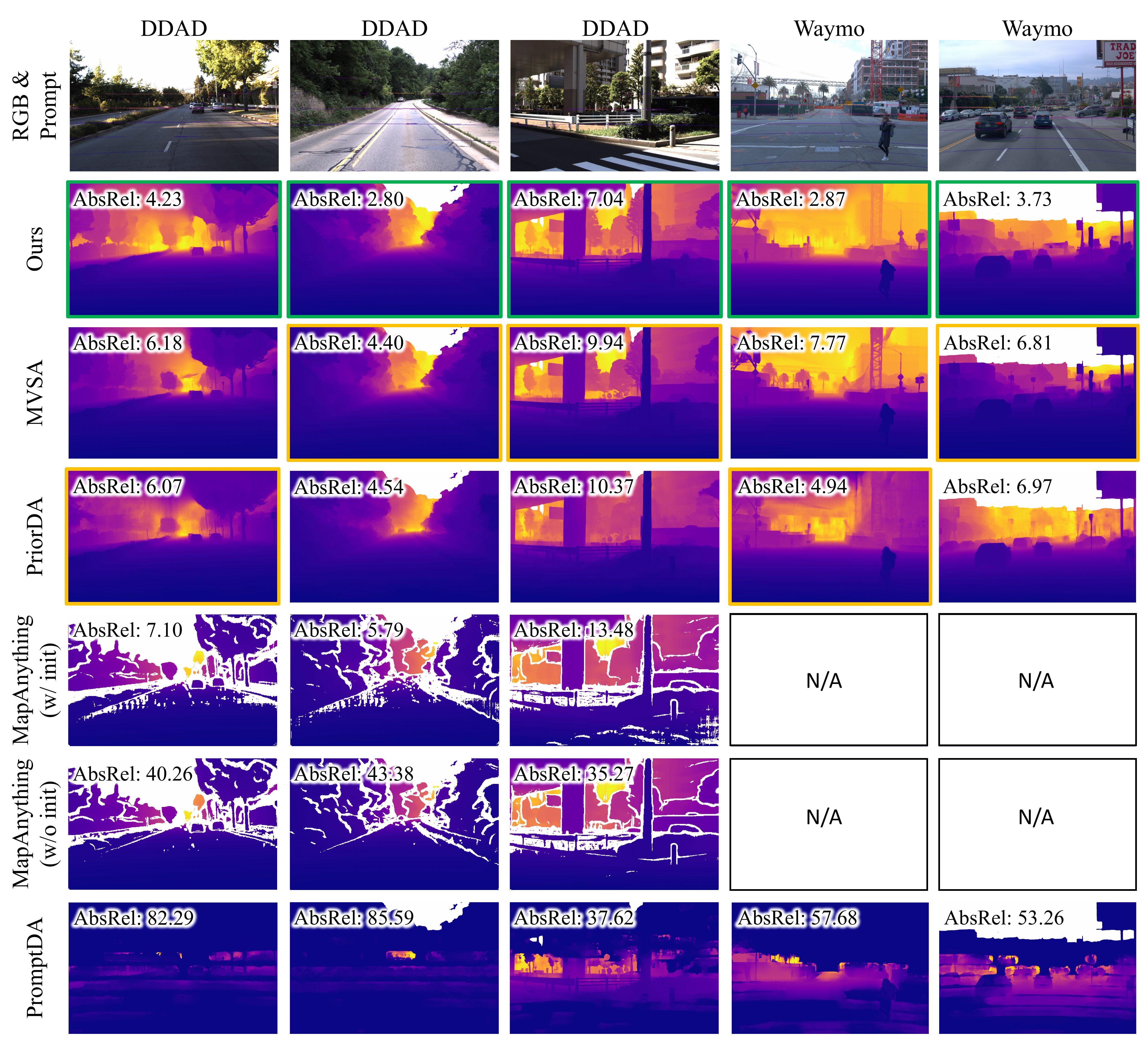}
    \end{center}
    \vspace{-0.2cm}
    \caption{
    \textbf{Qualitative comparison of depth prediction results on \textit{DDAD}~\cite{packnet2020ddad} and \textit{Waymo}~\cite{sun2020waymo}}.
    Rows show different methods: our DriveMVS model, MVSA~\cite{izquierdo2025mvsanywhere}, PriorDA~\cite{wang2025priorda}, MapAnything~\cite{keetha2025mapanything}, and PromptDA~\cite{lin2025promptda}, along with RGB and prompt inputs ($\boldsymbol{I_r}$, $\boldsymbol{P_r}$).
    For left to right: \textit{000156}, \textit{000155}, \textit{000194}, \textit{1024360143612057520}, \textit{12866817684252793621}.
    The \textcolor{forestgreen}{best} and \textcolor{yellowyellow}{second best} are highlighted with \textcolor{forestgreen}{green} and \textcolor{yellowyellow}{yellow} borders, respectively.
    }
\label{fig:sup_depth_2}
\end{figure*}

\begin{figure*}[t]
    \begin{center}
    \includegraphics[width=\linewidth]{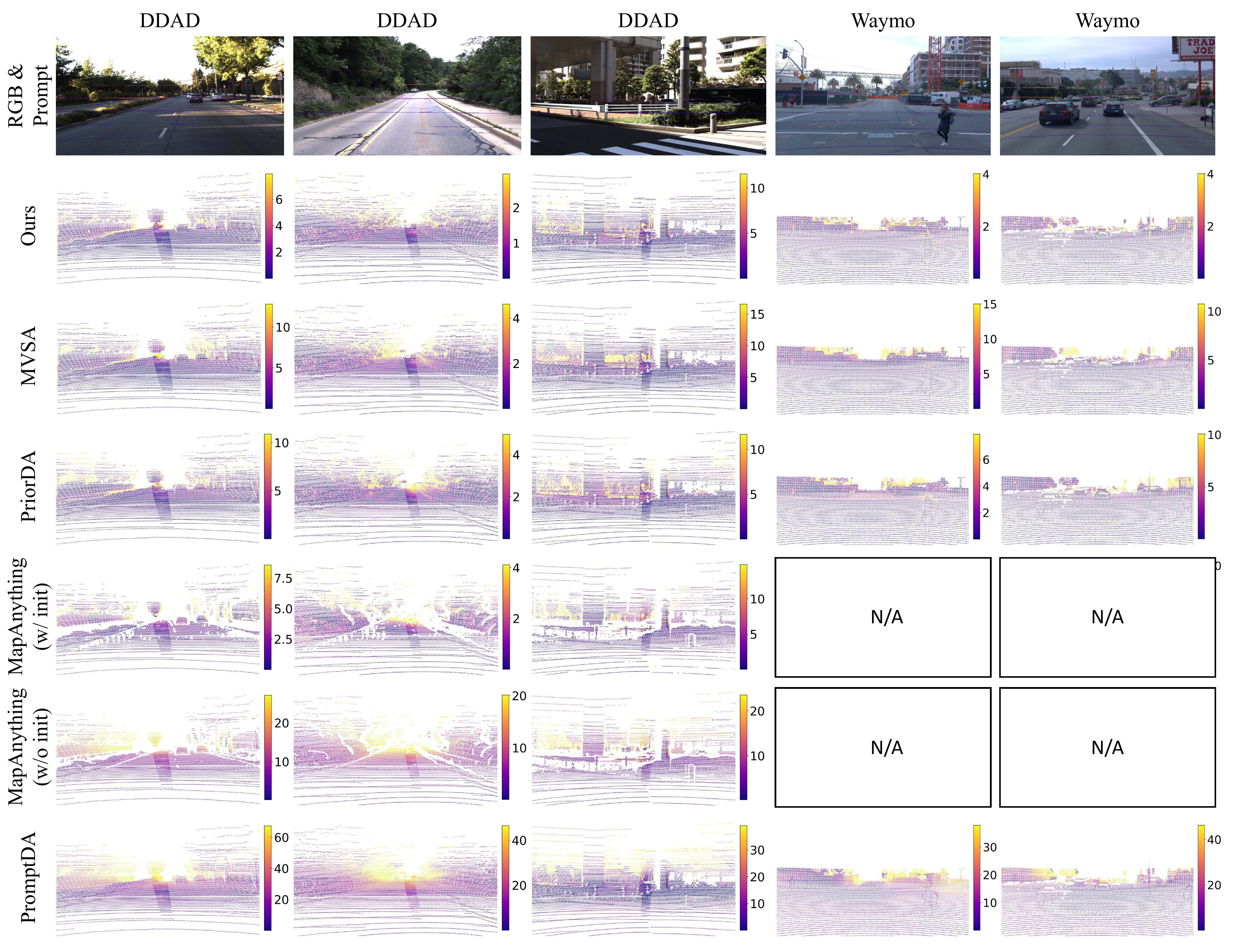}
    \end{center}
    \vspace{-0.2cm}
    \caption{
    \textbf{Qualitative comparison of depth prediction errors on \textit{DDAD}~\cite{geiger2012kitti} and \textit{Waymo}~\cite{sun2020waymo}}, corresponding to \cref{fig:sup_depth_2}.
    Rows show different methods: our DriveMVS model, MVSA~\cite{izquierdo2025mvsanywhere}, PriorDA~\cite{wang2025priorda}, MapAnything~\cite{keetha2025mapanything}, and PromptDA~\cite{lin2025promptda}, along with RGB inputs ($\boldsymbol I_r$) and prompt ($\boldsymbol P_r$).
    }
\label{fig:sup_error_2}
\end{figure*}

We present more visualization results of different methods across various datasets. \cref{fig:sup_depth_1} and \cref{fig:sup_error_1} show more cases on KITTI~\cite{geiger2012kitti} dataset. \cref{fig:sup_depth_2} and \cref{fig:sup_error_2} show more cases on DDAD~\cite{packnet2020ddad} and Waymo~\cite{sun2020waymo} dataset.

\begin{figure*}[t]
    \begin{center}
    \includegraphics[width=\linewidth]{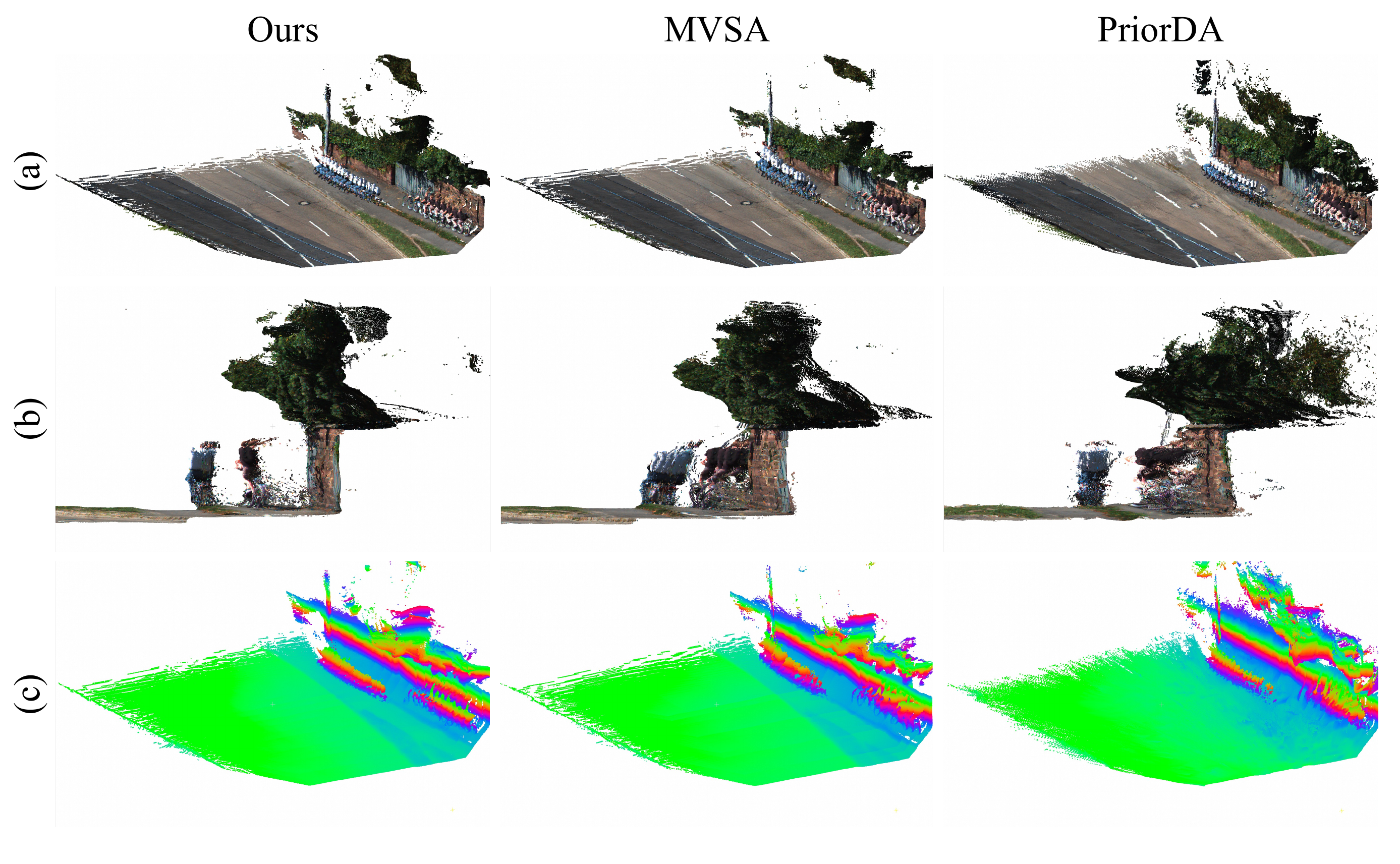}
    \end{center}
    \vspace{-0.2cm}
    \caption{
    \textbf{Qualitative comparison of 3D reconstruction results on \textit{KITTI}~\cite{geiger2012kitti}.}
    Rows show different methods: our DriveMVS model, MVSA~\cite{izquierdo2025mvsanywhere}, and PriorDA~\cite{wang2025priorda}.
    \textbf{(a)}: Reconstructed colored point clouds of the driving scene.
    \textbf{(b)}: Zoomed-in views highlighting fine-grained details of vegetation and structural boundaries.
    \textbf{(c)}: Corresponding $z$-axis visualizations.
    }
\label{fig:reconstruction}
\end{figure*}

\subsection{Detailed Analysis on More Experiments.}

\paragraph{Reconstruction Performance}
\cref{fig:reconstruction} presents a qualitative comparison on the KITTI dataset. 
As shown in rows (b) and (c), PriorDA struggles to maintain structural integrity in complex regions, exhibiting noticeable artifacts and floating noise around vegetation and walls.
While MVSA captures the general scene geometry, it tends to produce over-smoothed results with blurred boundaries. 
In contrast, ours effectively leverages the strengths of the MVS backbone. 
We achieve the most robust reconstruction with sharp boundaries and complete structural details, demonstrating superior performance in recovering both thin structures (\eg, tree trunks and pedestrians) and planar surfaces (\eg, road surface).
Specifically, we apply the same statistical outlier-removal filter to the reconstructed point clouds of all methods.

\begin{figure*}[t]
    \begin{center}
    \includegraphics[width=\linewidth]{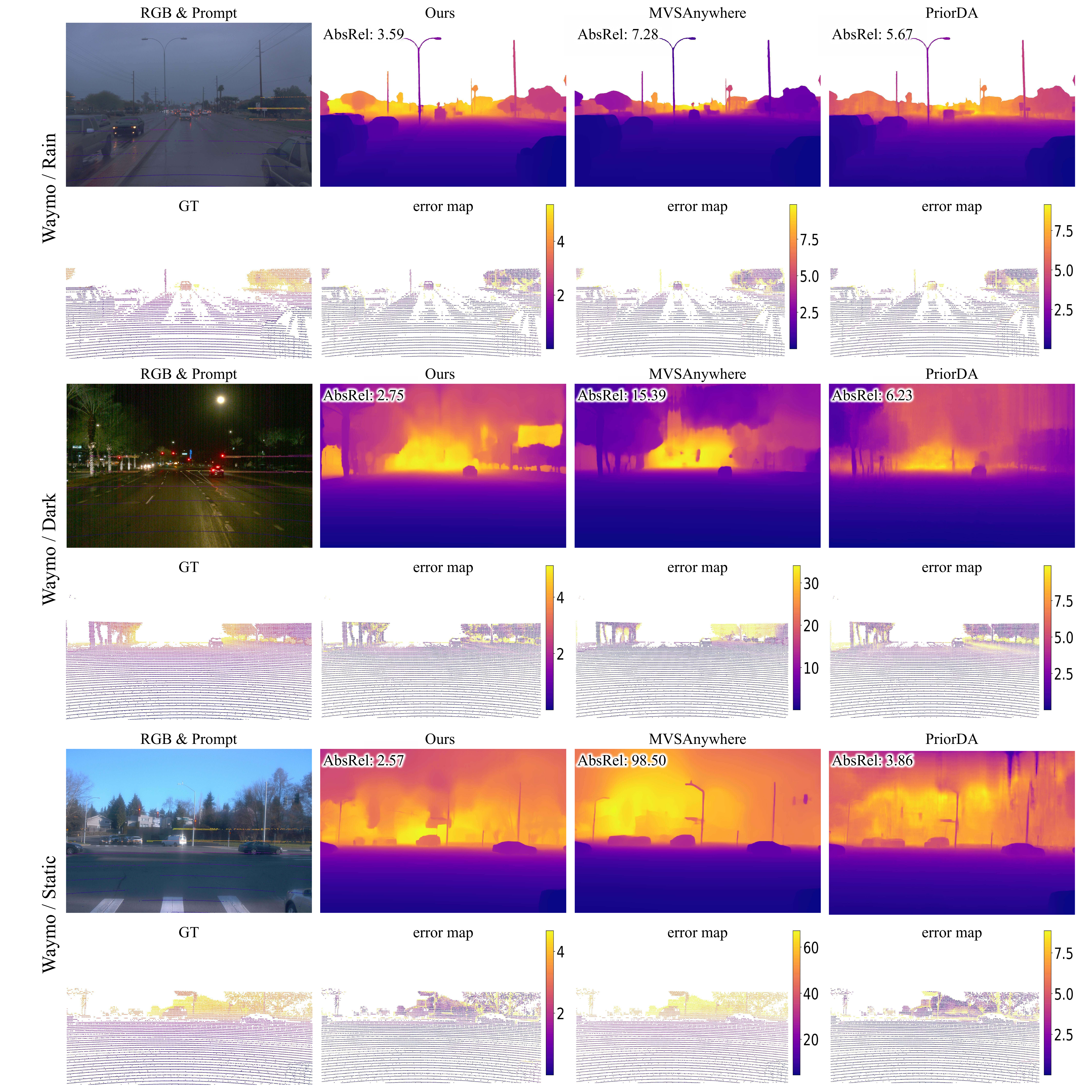}
    \end{center}
    \vspace{-0.2cm}
    \caption{
    \textbf{Qualitative comparison of depth prediction results and depth prediction errors across extreme cases for autonomous driving in our sampled dataset (most \textit{Waymo}~\cite{sun2020waymo}).}
    Columns show different methods: our DriveMVS model, MVSAnywhere~\cite{izquierdo2025mvsanywhere}, and PriorDA~\cite{wang2025priorda}, along with RGB \& prompt inputs ($\boldsymbol I_r,\boldsymbol P_r$) and ground-truth depths (GT).
    \textbf{[Top]}: \textit{11356601648124485814}.
    \textbf{[Middle]}: \textit{14107757919671295130}.
    \textbf{[Bottom]}: \textit{14127943474592757944}.
    }
\label{fig:sup_robustness}
\end{figure*}

\paragraph{Extreme Cases}
To rigorously evaluate model robustness under adverse conditions, we curated a specialized test set targeting three common yet challenging scenarios in autonomous driving: Rainy Weather, Low-light Environments, and Ego-static Situations (characterized by low-parallax motion). 
In \cref{sec:ablation}, we present a quantitative evaluation of these cases to demonstrate the stability of DriveMVS. 
Additionally, \cref{fig:sup_robustness} visualizes qualitative results sampled from the \textit{Waymo} and \textit{KITTI} datasets. 
Detailed dataset statistics, including specific scene IDs and the number of frames per scene, are provided in~\cref{tab:extreme_ids}.

\paragraph{Prompt Absence}
To assess the robustness of our model under partial sensor coverage, we conducted an experiment on LiDAR-blind view recovery, as illustrated in \cref{fig:multi-view}. 
In this setup, sparse LiDAR prompts are provided exclusively for the front-view camera, leaving the rear-view query image as a blind spot without direct geometric guidance.
As shown in the error maps, the baseline method (MVSAnywhere) fails to infer the correct metric scale for the rear view, resulting in an AbsRel error of 10.87\%. 
This indicates a limited capability in transferring geometric cues across disparate views.
In contrast, our method successfully leverages the spatial overlap and geometric correlations between views to propagate metric information from the prompted region (front) to the unprompted query view (rear). 
Consequently, our approach maintains accurate metric depth recovery (AbsRel: 6.83\%) even in the complete absence of current-view prompts, thereby verifying its effectiveness in handling sensor blind spots typical in autonomous driving.

\begin{table}
\centering
\caption{Speed and memory consumption.
Number benchmarked on an A100 GPU.
}
\scalebox{0.9}{%
\begin{tabular}{lcc}
\toprule
\textbf{\#} & \textbf{Time (ms)} & \textbf{Memory (GB)} \\
\midrule\midrule	
PriorDA & $246.79$ & $2.45$ \\
PromptDA & $300.60$ & $3.75$ \\
MVSA & $38.84$ & $4.02$ \\
Ours & $65.61$ & $4.47$ \\
\bottomrule
    \end{tabular}
    }

    \label{tab:consumption}
\end{table}

\subsection{Runtime Analysis}

We report the inference time cost and maximum GPU memory in \cref{tab:consumption}. 
Our framework strikes a favorable balance between performance and efficiency. 
Although slightly more resource-intensive than the lightweight MVSA due to enhanced feature aggregation, our method is significantly more efficient than prompt-guided methods such as PriorDA, demonstrating its potential for online autonomous driving applications.

% 作为补全算法的后续修正
% 训练过程中融入真实场景以及非zero-short的训练
% 不同prompt类型的影响 (VOID)
% 时序Window的影响


\end{document}